\documentclass[journal]{IEEEtran}
\usepackage{amsmath,amsfonts}

\usepackage{multirow}
\usepackage{slantsc}

\usepackage{booktabs}
\usepackage{threeparttable}
\usepackage{algorithm}
\usepackage{algpseudocode}
\usepackage{orcidlink}
\usepackage{array}
\usepackage[caption=false,font=normalsize,labelfont=sf,textfont=sf]{subfig}
\usepackage{textcomp}
\usepackage{stfloats}
\usepackage{url}
\usepackage{verbatim}
\usepackage{graphicx}
\def\BibTeX{{\rm B\kern-.05em{\sc i\kern-.025em b}\kern-.08em
    T\kern-.1667em\lower.7ex\hbox{E}\kern-.125emX}}
\usepackage{balance}
\usepackage{booktabs}
\usepackage{tabularx}

\usepackage{hyperref}
\hypersetup{
	colorlinks=true,
	linkcolor=blue,
	citecolor=blue,
	urlcolor=blue,
	pdfusetitle,      
	breaklinks=true      
}

\begin{document}
	
\title{SAMBA: A Scatter-Guided Masked Bidirectional Mamba Foundation Model for SAR Target Recognition}

\author{
Ke~Wang$^{\orcidlink{0009-0007-0994-1746}}$, \IEEEmembership{Graduate Student Member, IEEE}, 
Xiaoyi~Pan$^{\orcidlink{0000-0003-2175-4242}}$,
Zhaoyu~GU, 
Xiaofeng~Ai$^{\orcidlink{0000-0002-5080-8126}}$,
Zhiming~Xu$^{\orcidlink{0000-0002-3306-8023}}$,
Feng~Zhao$^{\orcidlink{0000-0003-1275-3352}}$, 
ShunPing~Xiao
	
\thanks{Manuscript received July 2026.
This work was supported in part by the National Natural Science Foundation of China under Grant Nos. 61701507, 61890542, and 61890540, the Youth Program of the National Natural Science Foundation of China under Grant No. 62401580, and in part by the Changsha Municipal Science and Technology Planning Project under Grant No. kq2209002.} 
\thanks{Ke Wang, Xiaoyi Pan, Zhaoyu Gu, Xiaofeng Ai, Zhiming Xu, Feng Zhao and Sunping Xiao are with the College of Electronic Science and Technology, National University of Defense Technology, Changsha 410073, China (E-mail: \href{mailto:wk\_wkk0310@163.com}{wk\_wkk0310@163.com}, \href{mrpanxy@nudt.edu.cn}{mrpanxy@nudt.edu.cn},
\href{guzhaoyu\_nudt@163.com}{guzhaoyu\_nudt@163.com}, \href{aixiaofeng@nudt.edu.cn}{aixiaofeng@nudt.edu.cn}, \href{zmxu\_nudt@163.com}{zmxu\_nudt@163.com}, \href{zhfbee@tom.com}{zhfbee@tom.com},  \href{xiaoshunping\_nudt@163.com}{xiaoshunping\_nudt@163.com})  (Corresponding author: Xiaoyi Pan)}
}

\markboth{Journal of \LaTeX\ Class Files,~Vol.~18, No.~9, September~2020}%
{How to Use the IEEEtran \LaTeX \ Templates}

\maketitle

\begin{abstract}
Synthetic aperture radar automatic target recognition (SAR ATR) plays a pivotal role in Earth observation and defense applications, yet its practical deployment is severely constrained by the scarcity of annotated training data. While self-supervised pre-training offers a promising solution to this label bottleneck, prevailing Transformer-based architectures suffer from prohibitively high quadratic computational complexity, and conventional universal masking strategies fail to account for the unique electromagnetic scattering properties inherent in SAR imagery. To address these limitations, we propose a scattering-guided bidirectional Mamba (SAMBA), an efficient self-supervised pre-training foundation model for SAR target interpretation. Our framework comprises three core technical innovations: (i) a linear-complexity Mamba encoder with a mid-sequence class token to mitigate computational bottlenecks; (ii) a three-level hierarchical Scattering-Guided Masked Autoencoder (SG-MAE) masking strategy guided by SAR physical priors, which aligns the pretext task with the intrinsic imaging mechanism of SAR data; and (iii) a lightweight SpatialMix feature interaction module to enhance cross-region feature fusion. Furthermore, we design a two-stage cross-domain pre-training pipeline to optimize the overall pre-training process. Extensive empirical evaluations demonstrate that SAMBA consistently delivers superior performance across all pre-training configurations, while featuring substantially fewer parameters than both convolutional neural network (CNN) and Transformer baselines. Compared with the default masking strategy in the standard MAE framework, the proposed SG-MAE masking strategy further boosts the model's few-shot transfer capability. Comprehensive benchmarking on seven downstream datasets covering both classification and detection tasks shows that SAMBA achieves state-of-the-art (SOTA) performance across most evaluation metrics, fully validating its robust generalizability across diverse SAR interpretation tasks. The source code and pre-trained weights of SAMBA are made publicly available at \url{https://github.com/mynswkk/SAMBA}.
\end{abstract}


\section{Introduction}
\IEEEPARstart{S}{ynthetic} Aperture Radar (SAR)~\cite{slesinskiReviewSyntheticAperture2025, yangPrincipalComponentMaximization2026, sunArbitrarydirectionSARShip2025a, langRecentAdvancesDeeplearningbased2025a} is an active microwave remote sensing modality and a core information acquisition tool in critical national strategic domains such as national defense and security, disaster monitoring, resource exploration, and maritime surveillance. Its irreplaceable status stems from its unique all-time, all-weather imaging capability.
As a core task of intelligent SAR image interpretation, SAR ATR~\cite{zhangCrosssensorSARImage2025, zhangLightweightSARShip2025a, zhouMaDiNetMambaDiffusion2025, chenTargetaspectDomainContinual2025} realizes automatic detection, localization, and identification of targets of interest in complex-background SAR imagery. It is widely applied in battlefield situational awareness, maritime traffic management, urban infrastructure monitoring, and disaster emergency response, serving as a key technical enabler for transforming raw SAR data into actionable knowledge.

In recent years, with the global launch and constellation-based operation of high-resolution SAR satellites, such as Gaofen-3, Sentinel-1, and TerraSAR-X, the volume of globally available SAR data has grown exponentially~\cite{huDistributedSpaceborneSAR2025, zhouDiffDet4SARDiffusionbasedAircraft2024a, wangTargetRecognitionSinglechannel2022}. The accumulation of massive unlabeled SAR data enables advances in data-driven intelligent interpretation, and offers new pathways to address the generalization bottleneck of traditional SAR ATR in complex scenarios.

\begin{figure}[htb]
	\centering
	\includegraphics[width=3.5in]{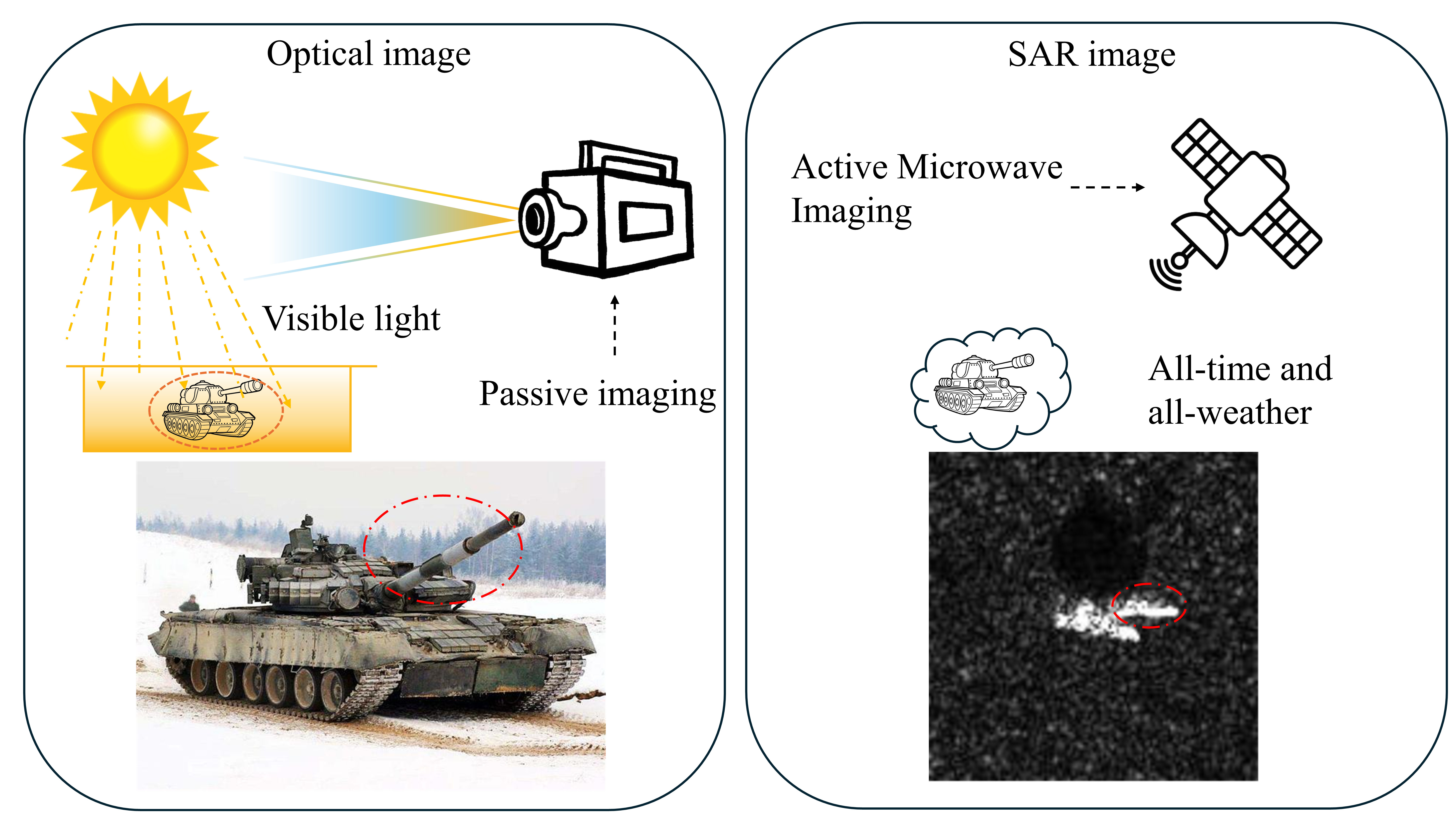}
	\caption{Imaging Principles of Optical Images and SAR Images.}
	\label{Fig:1}
\end{figure}

However, SAR and optical images differ fundamentally in imaging mechanisms, as illustrated in \figurename~\ref{Fig:1}. SAR acquires geospatial information via coherent imaging principles, with pixel intensities corresponding to the radar backscatter coefficients of ground targets. Consequently, SAR images inherently exhibit severe speckle noise, geometric distortion, and incidence-angle-dependent anisotropic scattering~\cite{vitaleSARDespecklingUsing2023, yangRobustOnestageDetector2022, zhangScatteringpointguidedRPNOriented2023, linUnpairedSpeckleExtraction2023, baiConditionalDiffusionSAR2024}.
These intrinsic properties make SAR image annotation highly resource-intensive. It requires professional expertise in radar remote sensing to distinguish targets from clutter and identify targets across diverse poses and configurations, along with substantial manual and material inputs, resulting in high costs and long lead times for constructing high-quality annotated datasets~\cite{wangLimiteddataSARATR2025a, wangNovelCFARbasedShip2024, fanSentinel1SARbasedGlobal2025, tsokasSARDataApplications2022}. Against this backdrop, mining intrinsic patterns from massive unlabeled SAR data and developing an efficient self-supervised pre-training framework tailored to SAR’s unique properties have emerged as a critical and pressing scientific challenge in intelligent SAR image interpretation.

To address the contradiction between explosive SAR data growth and extreme annotation scarcity, the field has undergone a distinct strategy evolution over the past decade: from the dominance of supervised learning, to the exploration of unsupervised learning, and ultimately to the rise of self-supervised learning (SSL). Each method shift has substantially advanced the capabilities of intelligent SAR image interpretation.

\textbf{The supervised learning era.} 
In 2016, Chen et al. first introduced deep convolutional neural networks (CNNs) into SAR ATR, inaugurating the deep learning era for intelligent SAR image interpretation~\cite{chenTargetClassificationUsing2016}. Since then, general-purpose visual backbones (e.g., ResNet, DenseNet, Swin Transformer) have been widely adapted to core SAR tasks including image classification, object detection, and semantic segmentation, and significantly outperform traditional machine learning algorithms on standard benchmarks such as MSTAR, SAR-Ship, and SSDD~\cite{weiHRSIDHighresolutionSAR2020, wangSARDatasetShip2019, huangWhatWhereHow2020, linSqueezeExcitationRank2019, xiaCRTransSarVisualTransformer2022a, dingConvolutionalNeuralNetwork2016, zhangLSSSDDv10DeepLearning2020}.

Nevertheless, the supervised learning method suffers from inherent critical limitations: it relies heavily on high-quality annotated data, yields models with limited generalization capacity, and fails to effectively exploit rich information embedded in massive unlabeled SAR data~\cite{duTargetDiscriminationBased2020, zhangDomainKnowledgePowered2022, yangActiveStylecontentDualbranch2026}.

\textbf{The unsupervised learning exploration phase.} 
To eliminate reliance on manually annotated data, researchers have investigated the application of conventional unsupervised learning in the SAR domain.
In 2021, Saha et al.~\cite{sahaBuildingChangeDetection2021} achieved unsupervised SAR-to-optical cross-domain translation via CycleGAN using paired unlabeled data, and accomplished building change detection by integrating deep change vector analysis with fuzzy inference rules on extracted bitemporal features.
In 2025, Tu et al.~\cite{tuMambaUDAMambaUnsupervised2025} proposed a Mamba-based unsupervised domain adaptation ship detection model with pseudo-label optimization to mitigate performance degradation of SAR ship detectors caused by domain shift and annotation scarcity.

Although conventional unsupervised learning requires no annotated data, its performance lags far behind supervised methods. Constrained by inherent limitations including reliance on handcrafted features, unstable clustering, and limited representation capacity, it cannot effectively address the annotation scarcity challenge in the SAR domain.

\textbf{The rise of self-supervised learning.}
SSL methods generate pseudo-supervisory signals from unlabeled data via elaborately designed pretext tasks, and then train the backbone network in a supervised manner. This strategy integrates the strengths of both unsupervised and supervised learning~\cite{moliniSpeckle2VoidDeepSelfsupervised2022, liPredictingGradientBetter2024a, Mo_2026_CVPR}.
In 2022, the MAE proposed by He et al.~\cite{heMaskedAutoencodersAre2022} achieved groundbreaking performance on optical imagery. With its compact architecture and strong generalizable feature learning capacity, it has opened a new pathway for SSL research in the SAR domain. MAE learns generic feature representations by randomly masking and reconstructing the majority of image regions, which enables full utilization of massive unlabeled data. In 2025, Wang et al.~\cite{wangFeatureGuidedMasked2025} proposed FG-MAE, which pioneered the adaptation of MAE to the SAR imagery domain. This method adopts handcrafted HOG features rather than raw pixels as the reconstruction target, which alleviates the adverse impact of inherent SAR speckle noise on the pre-training process to a certain degree. 

While existing studies have made notable progress in reconstruction optimization, speckle noise mitigation, and dataset construction, advancing SAR SSL from single-task solutions toward general-purpose foundation models, current research still suffers from the following key limitations:

1) \textbf{Computational bottleneck of Transformer architectures limits applicability to high-resolution SAR imagery.} Current SAR foundation models are predominantly built upon the Transformer and its variants. The self-attention mechanism inherent to these architectures exhibits \(\mathcal{O}(n^2)\) computational complexity, which scales quadratically with sequence length. With next-generation SAR sensors delivering continuously improving spatial resolution, input sequence lengths expand drastically. This trend sharply drives up the computational overhead of large-scale pre-training.

2) \textbf{Fundamental mismatch between masking strategies and SAR physical imaging mechanisms.} 
Most existing MAE-based SAR SSL methods directly inherit random masking strategies devised for optical imagery. Owing to the coherent imaging mechanism, SAR discriminative target information concentrates in a sparse set of high-intensity scattering centers. Under canonical uniform random masking, critical scattering centers are frequently masked, hindering the model from learning core target scattering features, while considerable computational resources are wasted on reconstructing uninformative background regions, resulting in suboptimal pre-training efficiency.

To address the aforementioned limitations, this paper proposes a SSL pre-training method for SAR images based on SAMBA. The main contributions of this paper are summarized as follows.

\begin{itemize}
	\item{We propose a foundational model named SAMBA for SAR ATR. It effectively addresses the common issues of inherent quadratic computational complexity and heavy computational overhead in conventional Transformer backbones.}
	\item{We present SG-MAE. Different from the pixel-wise random masking strategy widely used for natural images, this method adopts a three-level hierarchical masking strategy, including scatter point probabilistic masking, scale block masking and regional density adaptive regulation.}
	\item{We develop a lightweight spatial feature interaction module dubbed SpatialMix. It avoids the quadratic-complexity self-attention mechanism, uses 1D convolution (Conv1d) to implement efficient pixel-level spatial interaction, and achieves cross-dimensional channel feature fusion with channel-wise Multi-Layer Perceptron (MLP).}
\end{itemize}

\section{Related Work}

Driven by the rapid advancement of large-scale models, pre-training general-purpose foundation models on massive unlabeled SAR data has emerged as one of the most promising avenues to overcome prevailing technical bottlenecks. To this end, this work focuses on foundation models and pre-training strategies tailored for SAR ATR, and the following sections review representative studies in related fields.
\subsection{Model Architectures for SAR SSL}
The design of backbone architectures fundamentally dictates the performance upper bound of feature representation learning. For SAR self-supervised learning, backbone structures have undergone iterative evolution from CNNs to generic Vision Transformers (ViTs), as summarized in \tablename~\ref{tab:1}. The core motivation driving this evolution has consistently centered on enhancing the capacity to accurately model the unique physical characteristics of SAR imagery.

\begin{table*}[htb]
	\centering
	\caption{Summary of representative SAR ATR self-supervised learning methods. CL: Contrastive Learning, HPT: Hand-crafted Pretext Task, Masked IM: Masked Image Modeling}
	\begin{tabularx}{\linewidth}{l c c c c X X}
		\toprule
		\multicolumn{1}{c}{\textbf{Method}} 
		& \multicolumn{1}{c}{\textbf{Year}} 
		& \multicolumn{1}{c}{\textbf{Backbone}} 
		& \multicolumn{1}{c}{\textbf{Datasets}} 
		& \multicolumn{1}{c}{\textbf{Strategy}} 
		& \multicolumn{1}{c}{\textbf{Tasks}} 
		& \multicolumn{1}{c}{\textbf{Contribution}} \\
		\midrule
		CL-CNN~\cite{peiSelfsupervisedFeatureRepresentation2023} 
		& 2023 & CNN & MSTAR & CL & Target classification 
		& Contrastive learning pre-train CNN for two-stage SAR target classification. \\
		\addlinespace
		CL-ResNet50\cite{liuSelfsupervisedContrastiveLearning2023} 
		& 2023 & ResNet50 & MSTAR & CL & Target classification 
		& ResNet50 contrastive learning for cross-augmented SAR recognition. \\
		\addlinespace
		DCA-ResNet18\cite{zhaiDualConsistencyAlignment2023} 
		& 2023 & ResNet18 & MSTAR & CL & Target classification 
		& DCA uses pseudo-label alignment and FA to reduce SAR speckle noise. \\
		\addlinespace
		CFANet\cite{huangConvolutionalFeatureAggregation2024a} 
		& 2024 & CNN & MSTAR & HPT & Target classification 
		& CFANet aggregates features and uses pretext task for pre-training and sparse tuning. \\
		\addlinespace
		SAR-JEPA\cite{liPredictingGradientBetter2024a} 
		& 2024 & ViT-B & Multiple SAR & Masked & Target classification 
		& Leverages patch masking to perform SSL and MIM on noisy SAR imagery. \\
		\addlinespace
		SARATR-X\cite{liSARATRXBuildingFoundation2025a} 
		& 2025 & HiViT-B & ImageNet  \& Multiple SAR & Masked 
		& Target classification \& object detection
		& A foundation model for SAR target recognition across diverse tasks. \\
		\addlinespace
		SUMMIT\cite{duSUMMITSARFoundation2025} 
		& 2025 & ViT-B & Multi-SAR & HPT 
		& Target classification
		& Multi-auxiliary tasks enhance SAR features via SUMMIT and MuSID. \\
		\addlinespace
		\textbf{SAMBA} 
		& 2026 & Mamba & ImageNet  \& Multiple SAR & Masked 
		& Target classification and object detection 
		& A  first linear-time Mamba-based foundation model for SAR target recognition. \\
		\bottomrule
	\end{tabularx}
	\label{tab:1}
\end{table*}

Early SAR SSL research was dominated by contrastive learning frameworks, which generally adopted ResNet-series CNNs as backbone networks~\cite{peiSelfsupervisedFeatureRepresentation2023, huangConvolutionalFeatureAggregation2024a}. Representative works in this phase focused on the domain adaptation of generic contrastive learning and task-specific optimizations:
Liu et al.~\cite{liuSelfsupervisedContrastiveLearning2023} adapted the BYOL self-supervised contrastive learning framework to the SAR target recognition task with ResNet50 as the backbone, achieving performance significantly superior to that of traditional fully supervised methods under few-shot settings.
Zhai et al.~\cite{zhaiDualConsistencyAlignment2023} proposed the DCA-SSL model with ResNet18 as the backbone, which effectively mitigates the interference caused by SAR speckle noise and maintains favorable recognition accuracy even when the training set is noise-free while the test set is corrupted by random intensity noise.

However, CNNs are inherently limited by their local receptive fields~\cite{wangAdvSTMambaLightweightSpatial2026, qinRDBDINOImprovedEndtoend2025, liAirTargetIntent2025, wangNovelAutomatedNeural2025}, making it difficult for them to model long-range spatial dependencies in SAR images. The inductive bias of translation invariance in CNNs is also fundamentally incompatible with speckle noise and slant-range projection distortions in SAR imagery, leading to insufficient robustness of learned features~\cite{fangFEVTSARMulticategoryOriented2025}. To address this limitation, researchers have turned to Transformer architectures with global context modeling capabilities.
Deng et al.~\cite{dengSARImageRecognition2024} proposed a ViT-based contrastive SSL framework for SAR ATR, which yielded promising few-shot classification performance on the MSTAR dataset through large-scale unlabeled pre-training. Ma et al.~\cite{maSARViTVisionTransformer2024} pre-trained a ViT on large-scale unlabeled SAR data and fine-tuned it for downstream tasks to investigate the feasibility of Transformer models for SAR image analysis.
Subsequently, lightweight vision architectures such as LAD-Transformer and FastViT have also been widely adopted in the SAR community, which has alleviated the computational overhead to a certain extent~\cite{liuNovelLightweightAttentiondiscarding2023, ranFastViTRealtimeLinear2025}.
  
Nevertheless, the inherent quadratic self-attention complexity remains a fundamental bottleneck for Transformer-based SAR ATR. It drives prohibitive pre-training costs for large-scale datasets and prevents effective discriminative feature extraction from long-sequence high-resolution SAR imagery, severely hindering the advancement of large-scale high-resolution SAR foundation models.

\subsection{MAE-based Self-Supervised Training Strategy for SAR}
As masked image modeling (MIM) represented by MAE has emerged as the dominant self-supervised pre-training strategy in computer vision, MAE-based training strategies have also become a mainstream technical approach for constructing SAR foundation models~\cite{linSSMAESpatialSpectral2023, guoSCIIENetSharedComplementary2025a}.
Li et al.~\cite{liSARATRXBuildingFoundation2025a} developed SARATR-X, the first dedicated foundation model for SAR ATR. Equipped with HiViT as its backbone, this model effectively preserves high-resolution spatial features. The authors further designed a two-stage self-supervised pre-training pipeline that employs multi-scale gradient features as guidance signals to perform masked image modeling on SAR data, which effectively suppresses multiplicative speckle noise.
Du et al.~\cite{duSUMMITSARFoundation2025} proposed SUMMIT, a foundation model tailored for the SAR domain. Built upon a Transformer backbone, this model establishes a multi-auxiliary-task masked image modeling framework integrated with denoising and scattering feature enhancement mechanisms, and achieves strong performance across three downstream tasks: image classification, object detection, and instance segmentation.

Training strategy fundamentally determine the pre-training efficiency and representation quality of MAE-based frameworks\cite{jiResearchHeterogeneousRemote2025, caglayanSARWMixMAESARFoundation2025, wanMSPMAEMultiscalePerceptive2026}. Owing to the low signal-to-noise ratio (SNR), pronounced scattering heterogeneity, and inherent geometric distortion of SAR imagery, the direct transfer of optical MAE training strategies leads to imbalanced pre-training task difficulty, slow convergence, and inferior representational performance.

\section{METHODOLOGY}
\begin{figure*}[htb]
	\centering
	\includegraphics[width=7.25in]{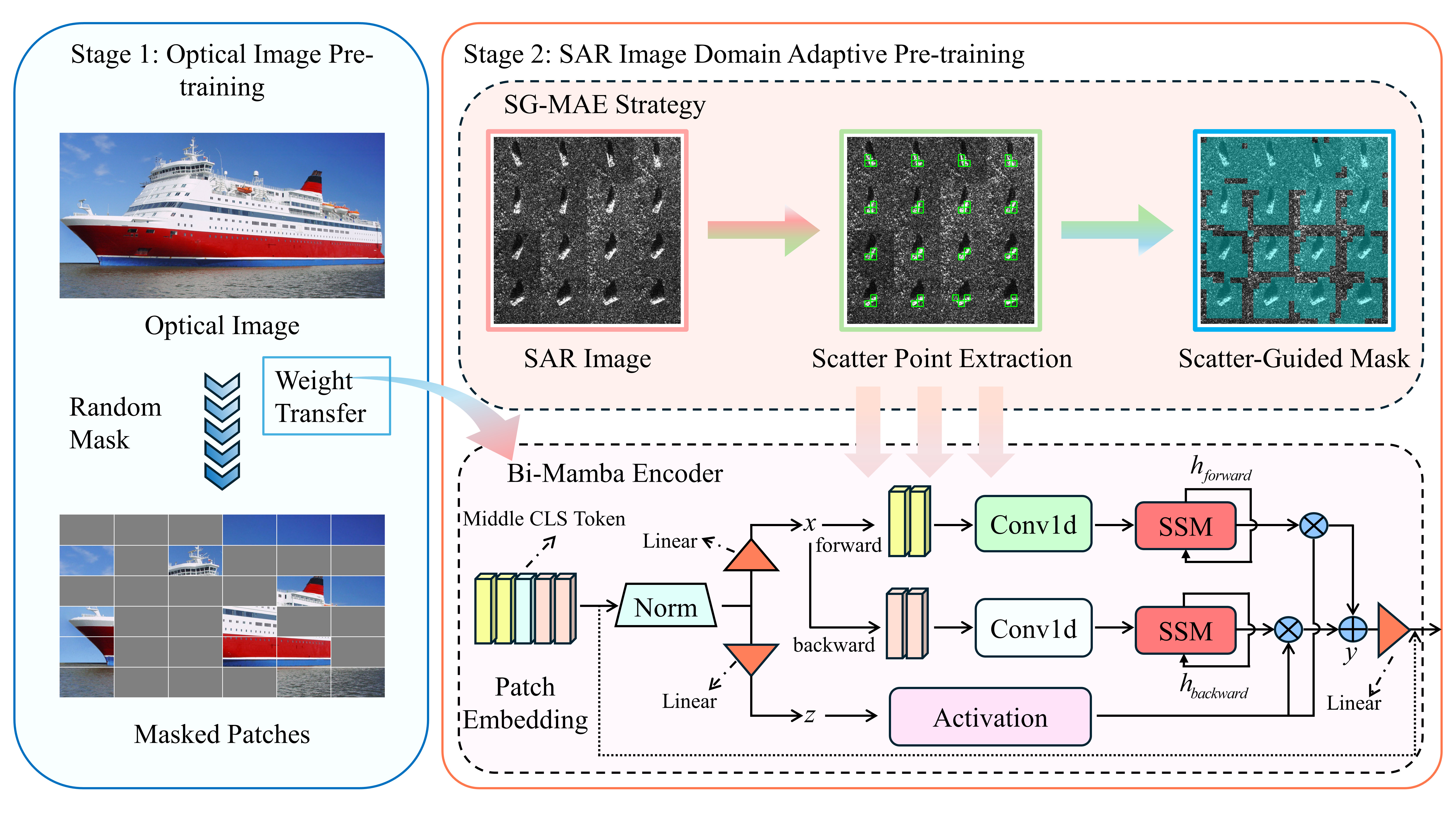}
	\caption{The overall framework of the proposed SAMBA method.}
	\label{Fig:2}
\end{figure*}

To address the aforementioned issues, this chapter proposes the SAMBA framework. First, we design an overlapping patch embedding module and a Mamba encoder backbone based on the Selective State Space Model (SSM)~\cite{guMambaLineartimeSequence2023}, and adopt an optimized layout with a middle-positioned class token (CLS)~\cite{pmlr-v235-zhu24f} to achieve global SAR feature modeling under linear time complexity. Second, leveraging the physical prior of SAR scattering imaging, we develop a three-level hierarchical masking strategy. Finally, we propose a lightweight SpatialMix decoder for efficient reconstruction of masked features. The overall architecture of the proposed model is shown in \figurename~\ref{Fig:2}.

\subsection{Mamba Encoder Backbone}
While the proposed Mamba encoder takes SAR images as input and outputs global semantic feature vectors, it consists of an overlapped patch embedding module and a bidirectional Mamba (Bi-Mamba) feature modeling module.

\textbf{Overlapping Patch Embedding.} 
Standard non-overlapped patch tokenization in visual backbones simplifies sequence construction but breaks local spatial continuity; for SAR imagery where semantics concentrate in scattering points, edges and extended targets, such hard partitioning fragments continuous targets and degrades boundary feature quality. To address this issue, we propose a 2D convolution-based overlapped patch embedding module that projects SAR images into 1D tokens with overlapping regions, enabling finer local feature capture and alleviating boundary information fragmentation.

Given an input SAR image $I\in\mathbb{R}^{H\times W\times C}$ ($C=1$ for single-channel SAR data), patch projection is implemented by a convolutional layer with kernel size $P\times P$ and stride $S$ ($S<P$). A stride smaller than the patch size ensures overlapping regions, preserving the continuity of SAR scattering structures across patch boundaries. This layer maps each overlapped patch into a $D$-dimensional feature vector, and the length of the generated token sequence is:
\begin{equation}
	M = \left\lfloor \frac{H-P}{S} + 1 \right\rfloor \times \left\lfloor \frac{W-P}{S} + 1 \right\rfloor
\end{equation}

Compared with non-overlapped linear projection, this convolution-based embedding incurs negligible additional computation while markedly enhancing tokens' local spatial perception, making it better suited for SAR data with fine-grained scattering features.

\textbf{Bi-Mamba Backbone.} 
Vanilla Mamba is tailored for autoregressive language modeling with only unidirectional sequence scanning, which is insufficient to capture omnidirectional spatial dependencies in 2D images. We therefore adopt a bidirectional architecture where each block contains two parallel SSM branches for forward and reverse sequence scanning.

The SSM computes output at position $t$ via hidden state recurrence:
\begin{equation}
	\begin{cases}
		\boldsymbol{h}_t = \boldsymbol{A} \boldsymbol{h}_{t-1} + \boldsymbol{B} \boldsymbol{x}_t \\
		y_t = \boldsymbol{C} \boldsymbol{h}_t
	\end{cases}
\end{equation}
with $\boldsymbol{h}_t \in \mathbb{R}^N$ as the hidden state and $\boldsymbol{A}, \boldsymbol{B}, \boldsymbol{C}$ as discretized parameters. Distinct from static SSMs, the SSM renders the discretization time step $\Delta$ input-dependent through linear projection with softplus activation for positivity:
\begin{equation}
	\Delta = \text{softplus}(\boldsymbol{W}_\Delta \boldsymbol{x} + b_\Delta)
\end{equation}
where $\boldsymbol{W}_\Delta$ and $b_\Delta$ denote the weight and bias of the linear projection layer. This selective mechanism enables adaptive information propagation and forgetting for improved contextual modeling.

The full pipeline of a single bidirectional Mamba block is as follows:
The input token sequence $\boldsymbol{T}_{\text{in}} \in \mathbb{R}^{B \times M \times D}$ first undergoes Layer Normalization, then is projected into two parallel branches via independent linear layers: the SSM branch $\boldsymbol{x} \in \mathbb{R}^{B \times M \times E}$ and the gating branch $\boldsymbol{z} \in \mathbb{R}^{B \times M \times E}$, where $E$ is the expanded hidden dimension.

In each bidirectional Mamba block, the input sequence is first layer-normalized and split into an SSM branch and a gating branch via independent linear projections. The SSM branch employs a small-kernel 1D depth-wise convolution to capture local dependencies, compensating for the weakness of pure global modeling in extracting fine-grained SAR scattering features. The enhanced features are then fed into forward and reverse SSM layers, whose outputs are summed to fuse bidirectional context:
\begin{equation}
	\boldsymbol{h}_{\text{bi}} = \boldsymbol{h}_{\text{fwd}} + \boldsymbol{h}_{\text{bwd}}
\end{equation}
where $\boldsymbol{h}_{\text{fwd}}$ and $\boldsymbol{h}_{\text{bwd}}$ represent outputs of the forward and reverse SSM layers, respectively.

The fused features are element-wise multiplied with SiLU-activated gating features, projected back to the original dimension, and added to the input via a residual connection to yield the block output:
\begin{equation}
	\boldsymbol{T}_{\text{out}} = \boldsymbol{T}_{\text{in}} + \boldsymbol{W}_o \left( \boldsymbol{h}_{\text{bi}} \odot \text{SiLU}(\boldsymbol{z}) \right)
\end{equation}
where $\odot$ denotes element-wise multiplication, and $\boldsymbol{W}_o \in \mathbb{R}^{E \times D}$ is the weight of the output projection layer. Moreover, the parallel scan algorithm enables efficient parallel SSM computation without explicit full kernel construction, preserving linear time complexity for long sequences.

\subsection{Scatter-Guided MAE }
\begin{figure}[htb]
	\centering
	\includegraphics[width=3.5in]{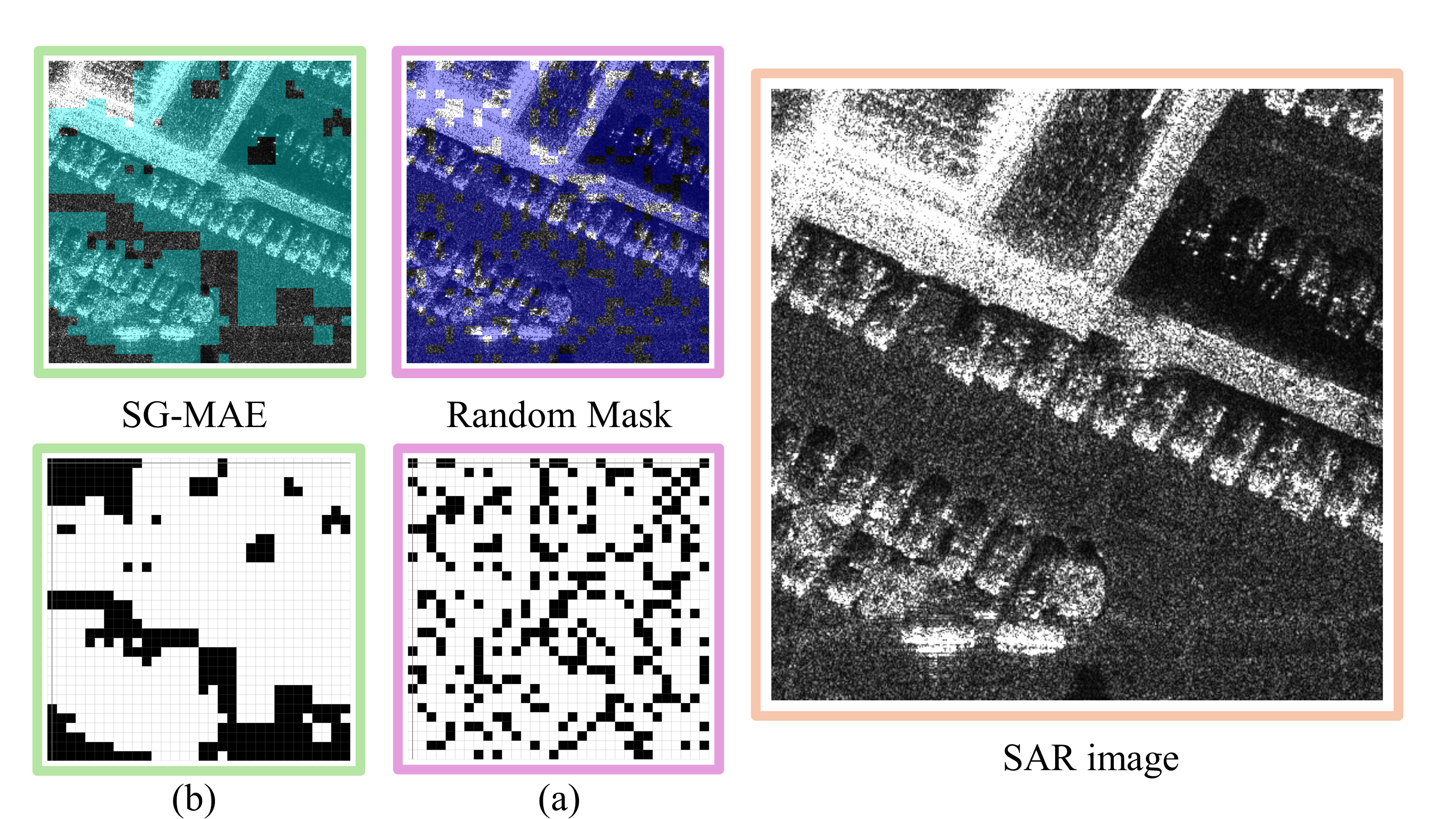}
	\caption{Comparison of different masking strategies in dense target SAR scenes. (a) The conventional random masking strategy, which causes excessive masking of sparse background regions; (b) The proposed SG-MAE masking strategy, which concentrates more on regions with target strong scatterers.}
	\label{Fig:3}
\end{figure}

Vanilla MAE with global uniform random masking fails to adapt to the information density disparity between scatterer-dense targets and sparse background regions, resulting in either insufficient target masking that weakens pre-training difficulty or excessive background masking that incurs redundant computational overhead, as illustrated in \figurename~\ref{Fig:3}. To address these limitations, we propose an SG-MAE strategy following a coarse-to-fine hierarchical guidance method, which progressively incorporates SAR physical priors at three granularity levels: adaptive regional density adjustment, multi-scale block structural constraint, and fine-grained scatterer contrast modulation.

\textbf{Level 3: Adaptive Regional Density Adjustment.}
As the coarsest-grained module for global mask allocation, this layer achieves region-adaptive masking: higher ratios for dense high-entropy target regions to promote structural reasoning, and lower ratios for sparse background regions to preserve reconstruction context.

It is implemented as follows: 1) A $5\times5$ sliding window traverses all SAR image patches, quantifying each central patch’s local density via scatterer counts within the window. 2) Per-sample max-normalization scales all local densities to $[0,1]$ using the in-image maximum $d_{\text{max}}$, producing $\text{density\_norm}$ and eliminating cross-image scatterer count variations.

The patch-wise mask ratio map $\text{ratio\_map}$ is derived from normalized density:
\begin{equation}
	\begin{aligned}
		\text{ratio\_map} &= \text{mask\_ratio} - \text{ratio\_delta} \\
		&\quad + 2 \cdot \text{ratio\_delta} \cdot \text{density\_norm}
		\label{eq:ratio_map}
	\end{aligned}
\end{equation}

where $\text{mask\_ratio}$ is the preset global target ratio, and $\text{ratio\_delta}$ controls adaptive ratio fluctuation, giving $\text{ratio\_map}$ a theoretical range of $[\text{mask\_ratio} - \text{ratio\_delta}, \text{mask\_ratio} + \text{ratio\_delta}]$. Finally, $\text{ratio\_map}$ is clipped to avoid fully masked or fully visible regions, ensuring pre-training stability.

\textbf{Level 2: Multi-scale Block Structural Constraint.}
Built upon the regional density adjustment, this medium-grained module strengthens scatterer spatial correlation modeling by discouraging trivial local interpolation.

Vanilla MAE masks individual patches, which fails to motivate long-range reasoning for SAR images with continuously distributed strong scatterers. We thus design a multi-scale block masking mechanism using irregular spatial blocks as basic masking units.

Specifically: 1) The input is divided into non-overlapping irregular blocks, with sizes randomly sampled from $1{\times}1$, $2{\times}2$ and $4{\times}4$ patches at probabilities of 0.3, 0.4 and 0.3, adapting to both isolated point scatterers and continuous planar structures. 2) All patches within a block share the same masking state: fully masked if the block is selected, otherwise fully visible.

This mechanism masks spatially adjacent scatterers in complete blocks, forcing the model to learn scatterer spatial distributions and structural correlations rather than simple interpolation, thus improving the structural representativeness of pre-trained features.

\textbf{Level 1: Scatterer Contrast Probabilistic Masking.}
As the finest-grained module, this layer further injects scatterer prior by finely calibrating patch-level masking probabilities to local information density.

In SAR images, core semantic information concentrates in strong scatterers and their adjacent structures, whereas backgrounds contain limited information. Complementing the regional density adjustment, patch-level scatterer features fine-tune masking probabilities to prioritize high-value scattering regions. Implementation details are provided in Algorithm~\ref{alg: 1}.

\begin{algorithm}[htbp]
	\caption{Scatterer Contrast Probabilistic Masking}
	\label{alg: 1}
	\begin{algorithmic}[1]
		\Require Original SAR image $I$, baseline ratio map $\text{ratio\_map}$,
		block set $\mathcal{B}$, target masking ratio $\text{mask\_ratio}$,
		coefficients $\alpha, \beta$
		\Ensure Final binary mask $\text{mask}$
		
		\Statex \textbf{Phase 1: Strong Scatterer Region Detection}
		\State Convert image to dB domain and perform speckle suppression filtering
		\State Extract high-brightness candidates via adaptive threshold $T = \mu + 1.2\sigma$
		\State Screen strong scatterers by local contrast and neighborhood bright-point constraints
		\State Generate binary scatterer label $\text{scatter\_label}$ and normalized contrast $\text{contrast\_norm}$
		
		\Statex \textbf{Phase 2: Fine-Grained Masking Probability Modulation}
		\State Compute patch-wise masking probability:
		$\text{prob}_i = \text{ratio\_map}_i + \alpha \cdot (2 \cdot \text{scatter\_label}_i - 1) + \beta \cdot \text{contrast\_norm}_i$
		\State Clip probabilities to $[0.02, 0.98]$ to avoid sampling failure
		
		\Statex \textbf{Phase 3: Final Sampling and Global Ratio Calibration}
		\State Aggregate patch probabilities to block-level probabilities $p_b$
		\State Generate initial mask via block-level Bernoulli sampling
		\State Calibrate global masking ratio to $\text{mask\_ratio}$ using Top-K strategy
		\State Output final calibrated mask $\text{mask}$
	\end{algorithmic}
\end{algorithm}

\subsection{SpatialMix Decoder}
The proposed SpatialMix decoder takes visible patch features output by the encoder, reconstructs features of masked regions via cross-patch interaction, and accomplishes full-image feature recovery. We replace the vanilla Transformer decoder in standard MAE with spatial mixing units composed of depthwise convolution and channel-wise MLP.

Each SpatialMix unit substitutes self-attention with spatial local interaction plus channel-wise global fusion, and adopts a Pre-LN dual-residual architecture for gradient stability. Given the input feature of the $l$-th unit $\boldsymbol{X}^{(l)} \in \mathbb{R}^{L \times D}$, where $L$ denotes sequence length and $D$ denotes feature dimension, the forward pass is formulated as:
 \begin{align}
 	\boldsymbol{X}_{\text{conv}}^{(l)} &= \boldsymbol{X}^{(l)} + \text{Conv1d}_{\text{dw}} \left( \text{LN} \left( \boldsymbol{X}^{(l)} \right) \right) \\
 	\boldsymbol{X}^{(l+1)} &= \boldsymbol{X}_{\text{conv}}^{(l)} + \text{MLP}_{\text{ch}} \left( \text{LN} \left( \boldsymbol{X}_{\text{conv}}^{(l)} \right) \right)
 \end{align}
 where $\text{LN}(\cdot)$ denotes layer normalization, $\text{Conv1d}_{\text{dw}}(\cdot)$ denotes depthwise Conv1d, and $\text{MLP}_{\text{ch}}(\cdot)$ denotes channel-wise MLP.
 
\textbf{ Depthwise Conv1d.}
 The spatial interaction module is implemented via depthwise Conv1d, with a kernel size of 3, padding of 1, and the number of groups equal to the feature dimension $D$. It performs local neighborhood convolution along the sequence dimension independently for each channel, modeling only spatial correlations between adjacent patches. The mathematical expression is:
 \begin{equation}
 	\boldsymbol{Y}_{:,d} = \boldsymbol{X}_{:,d} * \boldsymbol{k}_d, \quad d = 1,2,\dots,D
 \end{equation}
 where $\boldsymbol{X}_{:,d} \in \mathbb{R}^L$ is the feature of the $d$-th input channel, $\boldsymbol{k}_d \in \mathbb{R}^3$ is the convolution kernel for the corresponding channel, and $*$ represents the Conv1d operation. The computational complexity of this module is $O(L \cdot D)$, which scales linearly with the sequence length.
 
In SAR pre-training with a patch stride of 8, the token sequence can reach a scale of \(79 \times 79\). Standard self-attention with quadratic complexity \(O(L^2 \cdot D)\) incurs prohibitive computational and memory overhead. The linear complexity of depthwise convolution substantially reduces long-sequence decoding costs.
 
\textbf{Channel-wise Multi-layer Perceptron.}
 The channel-wise MLP adopts an expand-contract bottleneck structure, composed of two linear layers cascaded with a GELU activation function, which compensates for the lack of channel-wise interaction in depthwise convolution. Its formulation is:
 \begin{equation}
 	\text{MLP}_{\text{ch}}(\boldsymbol{X}) = \boldsymbol{W}_2 \cdot \text{GELU} \left( \boldsymbol{W}_1 \cdot \boldsymbol{X} + \boldsymbol{b}_1 \right) + \boldsymbol{b}_2
 \end{equation}
 where $\boldsymbol{W}_1 \in \mathbb{R}^{4D \times D}$ and $\boldsymbol{W}_2 \in \mathbb{R}^{D \times 4D}$ are the weight matrices of the two linear layers, and the feature dimension follows the transformation $D \to 4D \to D$. 
  
\section{Experiments}
\subsection{Experiment Dataset}
To validate the feature learning capacity and downstream generalization of the proposed pre-training framework, we adopt a two-stage pipeline: self-supervised pre-training on ImageNet to learn general visual priors, followed by self-supervised domain adaptation on a large unlabeled SAR dataset for feature alignment. Downstream performance is quantitatively evaluated on public SAR image classification and object detection benchmarks, with details presented as follows.

\textbf{Pre-training Datasets.}
Two datasets correspond to the two pre-training stages:
1) \textbf{ImageNet}\cite{dengImageNetLargescaleHierarchical2009}: Supports first-stage self-supervised pre-training to learn general low-level visual representations, providing well-initialized weights for subsequent SAR domain adaptation.
2) \textbf{186K SAR}\cite{liSARATRXBuildingFoundation2025a}: Contains 186,000 unlabeled SAR images covering diverse targets, frequency bands, depression angles and complex backgrounds.

\textbf{Classification Datasets.}
Three widely adopted SAR classification benchmarks are selected for comprehensive evaluation:
1) \textbf{MSTAR}: A standard SAR ATR benchmark with X-band images of 10 ground military target categories.
2) \textbf{FUSAR-Ship}\cite{houFUSARshipBuildingHighresolution2020}: A ship-specific SAR dataset covering various ship categories, marine environments, sensors and resolutions.
3) \textbf{SAR-ACD}\cite{sunSCANScatteringCharacteristics2022}: A fine-grained SAR classification dataset with abundant annotations under complex imaging conditions.

\textbf{Detection Datasets.}
Four public benchmarks are chosen for SAR object detection evaluation across different scales and scenarios:
1) \textbf{SSDD}\cite{zhangLSSSDDv10DeepLearning2020}: A classic SAR ship detection benchmark with manually annotated bounding boxes, spanning diverse sea states, imaging modes and resolutions.
2) \textbf{SARDet-100K}\cite{li2024sardet}: A large-scale detection dataset with hundreds of thousands of annotated SAR images, covering rich scenes, target scales and imaging conditions.
3) \textbf{SIVED}\cite{linSIVEDSARImage2023a}: A ground vehicle SAR dataset with multiple vehicle types and varied imaging parameters.
4) \textbf{SAR-Aircraft}\cite{wang2023saraircraft}: An aerial target SAR dataset consisting of multiple aircraft models.

\subsection{Comparison of Model Backbones and Pre-training Strategies}
To evaluate different backbones under various self-supervised pre-training schemes, we adopt the unified experimental pipeline, pre-training dataset, and linear probing protocol from \cite{liSARATRXBuildingFoundation2025a}. \tablename~\ref{tab:2} presents few-shot classification accuracies of ConvNeXt-V2 \cite{wooConvNeXtV2Codesigning2023}, ViT \cite{dosovitskiy2021image}, HiViT \cite{zhangHiViTHierarchicalVision2022}, and our Mamba-based SAMBA backbone across four setups: supervised ImageNet initialization, single-domain self-supervised pre-training on ImageNet and SAR data respectively, and two-step cross-domain pre-training. For fair comparison, results of ConvNeXt-V2, ViT and HiViT are reproduced from \cite{liSARATRXBuildingFoundation2025a}, while all Mamba-based experiments are newly implemented in this work.

\begin{table*}[htbp]
	\centering
	\renewcommand{\arraystretch}{1.3} 
	\setlength{\tabcolsep}{4pt}   
	\caption{Parameter comparison and classification accuracy (\%) of different backbones and pre-training strategies under 5/10/20-shot SAR target classification. Baseline results of ConvNeXt-V2, ViT and HiViT are reproduced from\cite{liSARATRXBuildingFoundation2025a}. Mamba-based comparative groups are supplemented in this work.}
	\label{tab:2}
	\begin{tabular*}{\linewidth}{@{\extracolsep{\fill}} c c l l l c c c @{}}
		\toprule
		\multirow{2}{*}{Backbones} & \multirow{2}{*}{Params} & \multicolumn{3}{c}{Pre-Training} & \multicolumn{3}{c}{Classification (N-Shot)} \\
		\cline{3-5} \cline{6-8}
		& & Settings & Dataset & Method & 5 & 10 & 20 \\
		\hline
		ConvNeXt-V2 & 89M & SL-ImageNet & ImageNet-1K & Supervised & 52.5 & 61.7 & 70.5 \\
		ConvNeXt-V2 & 89M & SSL-ImageNet & ImageNet-1K & FCMAE & 47.2 & 54.5 & 64.0 \\
		ConvNeXt-V2 & 89M & SSL-SAR & SAR images & FCMAE & 52.7 & 60.9 & 67.7 \\
		ConvNeXt-V2 & 89M & SSL-ImageNet \& SAR & ImageNet \& SAR & FCMAE & 54.7 & 61.5 & 69.5 \\
		\hline
		ViT & 86M & SL-ImageNet & ImageNet-1K & Supervised & 58.6 & 65.7 & 74.2 \\
		ViT & 86M & SSL-ImageNet & ImageNet-1K & MAE & 50.7 & 58.0 & 65.5 \\
		ViT & 86M & SSL-SAR & SAR images & MAE & 54.1 & 61.5 & 68.2 \\
		ViT & 86M & SSL-ImageNet \& SAR & ImageNet \& SAR & MAE & 65.8 & 76.4 & 83.6 \\
		\hline
		HiViT & 66M & SL-ImageNet & ImageNet-1K & Supervised & 49.0 & 55.8 & 63.3 \\
		HiViT & 66M & SSL-ImageNet & ImageNet-1K & MAE & 53.0 & 60.3 & 69.3 \\
		HiViT & 66M & SSL-SAR & SAR images & MAE & 64.9 & 72.7 & 79.9 \\
		HiViT & 66M & SSL-ImageNet \& SAR & ImageNet \& SAR & MAE & 71.5 & 78.5 & 84.0 \\
		HiViT & 66M & SSL-ImageNet \& SAR & ImageNet \& SAR & SARATR-X & \underline{76.5} & \underline{80.8} & \underline{85.1} \\
		\hline
		Mamba & 27M & SL-ImageNet & ImageNet-1K & Supervised & 54.6 & 61.3 & 67.8\\
		Mamba & 27M & SSL-ImageNet & ImageNet-1K & MAE & 61.9 & 68.4 & 70.88 \\
		Mamba & 27M & SSL-SAR & SAR images & MAE & 66.9 & 72.5 & 76.7\\
		Mamba & 27M & SSL-ImageNet \& SAR & ImageNet \& SAR & MAE & 72.9 & 78.9 &  82.1\\
		\textbf{Mamba} & 27M & SSL-ImageNet \& SAR & ImageNet \& SAR & \textbf{Ours (SG-MAE)} & \textbf{80.6} & \textbf{83.6} & \textbf{88.1} \\
		\bottomrule
		\multicolumn{5}{l}{\footnotesize \textbf{Bold} indicates the best performance; \underline{underline} indicates the second-best performance.}
	\end{tabular*}
\end{table*}

Horizontal comparisons across backbones yield consistent conclusions under all pre-training configurations, with clear gaps in classification performance and parameter efficiency.
ConvNeXt-V2 has the largest parameter count of 89M among compared models and achieves the lowest overall accuracy.
Vanilla ViT contains 86M parameters and adopts global self-attention. It substantially outperforms ConvNeXt-V2, but suffers from high computational cost and limited local texture modeling capacity for high-resolution SAR scenarios.
HiViT reduces parameters to 66M and adopts hierarchical window attention, striking a better balance between local scatterer detail perception and global structural modeling. It consistently outperforms vanilla ViT despite fewer parameters.
Our proposed Mamba-based method contains only 27M parameters, less than one-third of ConvNeXt-V2 and roughly 41\% of HiViT, yet achieves state-of-the-art performance under every pre-training scheme.
Benefiting from linear-complexity global sequence modeling, Mamba inherently adapts to the sparse distribution of SAR scattering features, extracting more discriminative target features at significantly lower computational and parameter costs.

Self-supervised pre-training solely on optical ImageNet suffers severe domain mismatch with SAR coherent imaging properties and yields the poorest downstream transfer performance, while direct SAR self-supervised pre-training effectively narrows the domain gap but is constrained by limited training samples. The two-step cross-domain pre-training strategy, which first transfers general low-level visual priors from large-scale ImageNet data and then refines scatterer-aware feature extraction on the SAR dataset, delivers superior performance. Under this two-step cross-domain setting, we further validate the effectiveness of the proposed SG-MAE three-level hierarchical masking strategy: compared with standard random masking in vanilla MAE, SG-MAE boosts 5/10/20-shot few-shot classification accuracy from 71.5/78.5/84.0 to 80.6/83.6/88.1, respectively.

\subsection{Experimental Analysis of Downstream Tasks}
To comprehensively assess the performance of the proposed pre-training framework on downstream SAR interpretation tasks, we conduct comparative experiments on multiple widely adopted benchmark datasets against representative methods from the literature.

\textbf{Classification task.}
We conduct few-shot fine-tuning evaluation on three representative benchmarks, namely MSTAR, FUSAR-Ship and SAR-ACD, and adopt experimental results from existing literature as comparison baselines. The quantitative comparison results across all three datasets are summarized in \tablename~\ref{tab:3}.

\begin{table}[htbp]
	\centering
	\begin{threeparttable}
		\caption{Few-Shot Classification Performance Comparison on Three SAR Datasets (\%)}
		\footnotesize
		\setlength{\tabcolsep}{4.0pt}
		\label{tab:3}
		\begin{tabular}{lcccc}
			\toprule
			\multicolumn{5}{c}{MSTAR (Standard-10way)} \\
			\midrule
			Method & Year & 1-shot & 2-shot & 5-shot \\
			\midrule
			DKTS-N\cite{zhangDomainKnowledgePowered2022} & 2021 & 49.3 & 58.5 & 72.3 \\
			ConvT\cite{wangGlobalLocalConvolutional2022} & 2022 & 42.6 & 54.4 & 75.2\\
			CRID\cite{wangCrucialFeatureCapture2023} & 2023 & 48.3 & 51.0 & 72.4 \\
			PD\cite{zhangOptimalAzimuthAngle2024} & 2024 & 46.7 & 58.9 & 90.3 \\
			SARATR-X\cite{liSARATRXBuildingFoundation2025a} & 2025 & \underline{85.2} & \underline{91.4} & \textbf{95.9} \\
			\midrule
			Ours & 2026& \textbf{89.6} (+4.4) & \textbf{93.5} (+2.1) & \underline{95.0} (-0.9) \\
			\midrule
			\multicolumn{5}{c}{FUSAR-Ship (10way)} \\
			\midrule
			Method & Year & 1-shot & 2-shot & 30\% \\
			\midrule
			ResNet-50\cite{heDeepResidualLearning2016a} & 2016 & - & - & 58.4  \\
			ViT\cite{dosovitskiy2021image} & 2021 & - & - & 71.1  \\
			Swin-T\cite{liuSwinTransformerHierarchical2021} & 2021& - & - & 60.31\\
			SUMMIT\cite{duSUMMITSARFoundation2025} & 2025 & - & - & \underline{71.9} \\
			\midrule
			Ours & 2026 & \textbf{80.0} & \textbf{81.8}  & \textbf{85.0} (+13.1) \\
			\midrule
			\multicolumn{5}{c}{SAR-ACD (5way)} \\
			\midrule
			Method & Year & 1-shot & 2-shot & 30\% \\
			\midrule
			ResNet-50\cite{heDeepResidualLearning2016a} & 2016 & - & - & 59.7  \\
			ViT\cite{dosovitskiy2021image} & 2021& - & -& 80.0\\
			Swin-T\cite{liuSwinTransformerHierarchical2021} & 2021 & - & - & 63.18 \\
			SUMMIT\cite{duSUMMITSARFoundation2025} & 2025& - & - & \textbf{84.3} \\
			\midrule
			Ours & 2026 & \textbf{55.9}  & \textbf{65.7} & \underline{81.1} (-3.2) \\
			\bottomrule
		\end{tabular}
		\begin{tablenotes}[flushleft]
			\footnotesize
			\item[] \textbf{Bold} indicates the best performance; \underline{underline} indicates the second-best performance.
		\end{tablenotes}
	\end{threeparttable}
\end{table}
Our proposed method achieves accuracy improvements over existing approaches on the MSTAR dataset, and establishes few-shot recognition performance on both FUSAR-Ship and SAR-ACD benchmarks. These results demonstrate the robust and stable recognition capability of the proposed model across diverse scenarios.

\textbf{Detection task.}
We evaluate the generalization capacity of the proposed method on four benchmark datasets: SSDD, SARDet-100K, SIVED, and SAR-Aircraft, which cover three typical detection scenarios: ship, ground vehicle, and aircraft targets.
Comparative results against existing detection methods are presented in \tablename~\ref{tab:4}.
\begin{table}[htbp]
	\centering
	\begin{threeparttable}
		\footnotesize
		\setlength{\tabcolsep}{4.0pt}
		\caption{Detection Performance Comparison on SAR Datasets (\%)}
		\label{tab:4}
		\begin{tabular}{lcccc}
			\toprule
			\multicolumn{5}{c}{SARDet-100K (Object Detection)} \\
			\midrule
			Method & Year & mAP & $\text{mAP}_{50}$ & $\text{mAP}_{75}$ \\
			\midrule
			Deformable DETR\cite{zhuDeformableDETRDeformable2020} & 2020 & 50.0 & 85.1 & 51.7 \\
			Swin Transformer\cite{liuSwinTransformerHierarchical2021} & 2021 & 53.8 & 87.8 & 59.0 \\
			ConvNeXt\cite{liuConvNet2020s2022a}  & 2022 & 55.1 & 87.8 & 59.5 \\
			MSFA\cite{li2024sardet} & 2024 & 56.4 & 88.2 & 61.5 \\
			SARATR-X\cite{liSARATRXBuildingFoundation2025a} & 2025 & \underline{57.3} & 88.7 & \underline{62.8} \\
			SUMMIT\cite{duSUMMITSARFoundation2025} & 2025& 57.0 & \textbf{89.9} & 62.7 \\
			\midrule
			Ours & 2026 & \textbf{59.7} (+2.4) & \underline{89.1} (-0.8) & \textbf{68.0} (+5.2) \\
			\midrule
			\multicolumn{5}{c}{SSDD (Ship Detection)} \\
			\midrule
			Method & Year & mAP & $\text{mAP}_{50}$ & $\text{mAP}_{75}$ \\
			\midrule
			FBR-Net\cite{fuAnchorfreeMethodBased2021} & 2021 & -- & 94.1 & 59.1 \\
			CRTransSar\cite{xiaCRTransSarVisualTransformer2022a} & 2022 & -- & 97.0 & 76.2 \\
			CS$^n$Net\cite{chenCSNetRemote2023} & 2023 & 64.9 & 97.1 & -- \\
			SARATR-X\cite{liSARATRXBuildingFoundation2025a} & 2025 & 67.5 & \underline{97.3} & 83.5 \\
			SUMMIT\cite{duSUMMITSARFoundation2025} & 2025& \underline{70.4} & 96.7 & \textbf{86.3} \\
			\midrule
			Ours & 2026 & \textbf{71.3} (+0.9) & \textbf{97.8} (+0.5) & \underline{85.6} (-0.7) \\
			\midrule
			\multicolumn{5}{c}{SIVED (Ground Vehicle Detection)} \\
			\midrule
			Method & Year & mAP & $\text{mAP}_{50}$ & $\text{mAP}_{75}$ \\
			\midrule
			Faster R-CNN\cite{yangRotatedFasterRCNN2020} & 2020 & 53.1 & 97.8 & 51.0\\
			Gliding Vertex\cite{xuGlidingVertexHorizontal2021}& 2021 & 56.1 &  98.3 & 59.6\\
			Oriented RepPoints\cite{linSIVEDSARImage2023a} & 2023 & \underline{60.2} & \underline{99.1} & \underline{70.7} \\
			\midrule
			Ours & 2026 & \textbf{67.6} (+7.4) & \textbf{97.3} (-1.8) & \textbf{82.5} (+11.8) \\
			\midrule
			\multicolumn{5}{c}{SAR-Aircraft (Aircraft Detection)} \\
			\midrule
			Method & Year & mAP & $\text{mAP}_{50}$ & $\text{mAP}_{75}$ \\
			\midrule
			RepPoints\cite{yangRepPointsPointSet2019} & 2019 & -- & 72.6 & 53.3 \\
			SKG-Net\cite{fuScatteringkeypointguidedNetworkOriented2021}& 2021 & -- & 70.7 & 46.4 \\
			SA-Net\cite{wang2023saraircraft} & 2023 & -- & 77.7 & 62.8 \\
			SARATR-X\cite{liSARATRXBuildingFoundation2025a} & 2025 & \underline{58.7} & \textbf{86.1} & \underline{64.7} \\
			SUMMIT\cite{duSUMMITSARFoundation2025} & 2025& 57.7 & 81.4 & 64.4 \\
			\midrule
			Ours & 2026 & \textbf{60.0} (+1.3) & \underline{85.0} (-1.1) & \textbf{66.5} (+1.8) \\
			\bottomrule
		\end{tabular}
		\begin{tablenotes}[flushleft]
			\footnotesize
			\item[] \textbf{Bold} indicates the best performance; \underline{underline} indicates the second-best performance.
		\end{tablenotes}
	\end{threeparttable}
\end{table}

Our proposed method achieves consistent improvements in detection performance over existing baselines across all four SAR object detection datasets. Furthermore, as these datasets cover distinct imaging scenarios, the results further validate the cross-scene generalization effectiveness of our approach.

Overall, the proposed method yields consistent performance improvements on most metrics across all classification and detection benchmarks, which substantiates the effectiveness and strong generalizability of the proposed pre-training framework for diverse SAR interpretation tasks.

\subsection{Visualization}
In this section, we visualize the computational efficiency and detection performance of the proposed SAMBA model.
 We conduct a systematic benchmark on six representative architectures, including ResNet-50, Swin-Base, ViT-Base, DeiT-Base, HiViT-Base, and the proposed SAMBA, to thoroughly verify the computational efficiency and scalability of our method from three complementary perspectives.

\begin{figure*}[t]
	\centering
	
	\subfloat[]{\includegraphics[width=0.32\textwidth]{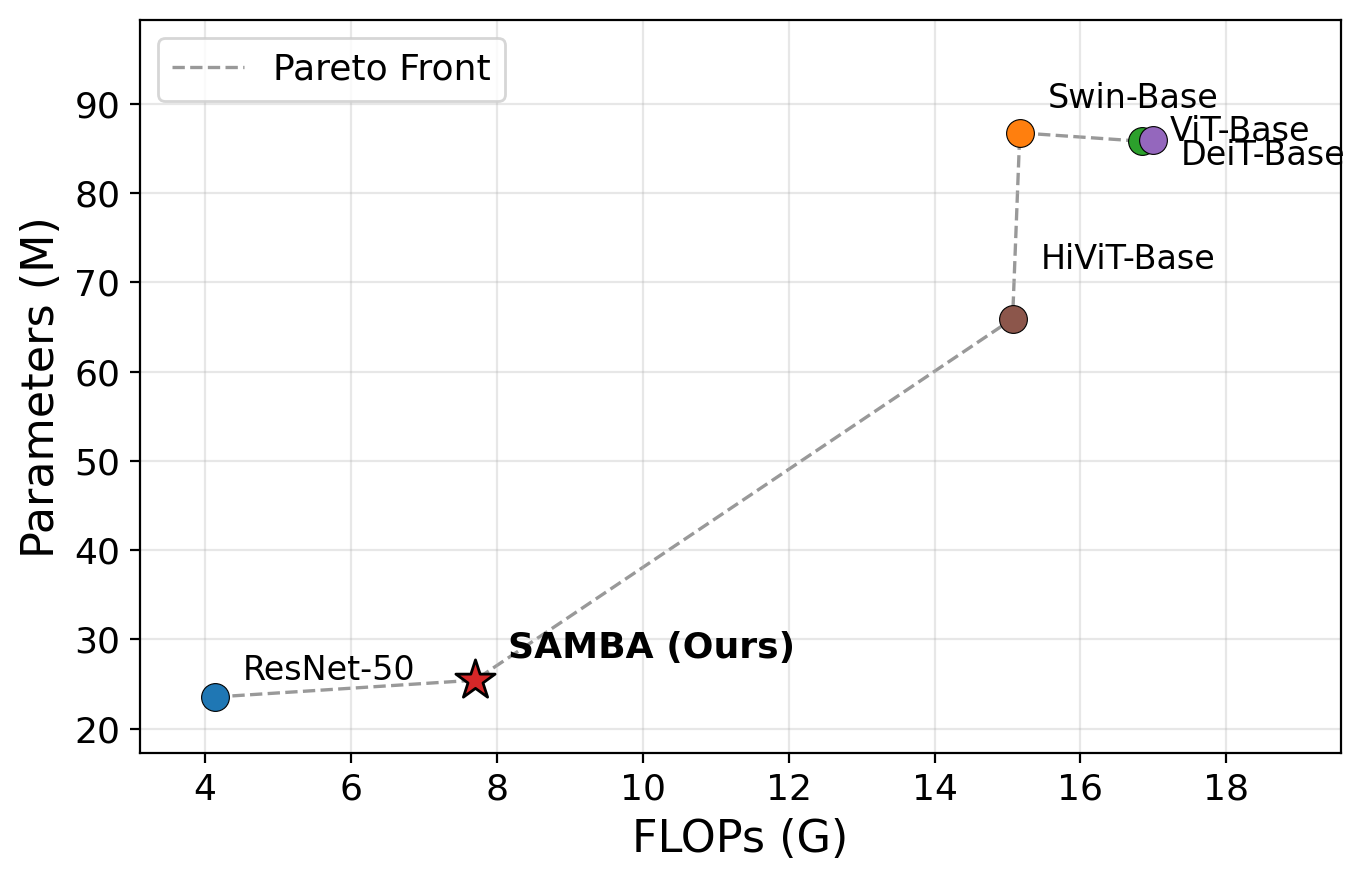}\label{subfig:a}}
	\hfill
	\subfloat[]{\includegraphics[width=0.32\textwidth]{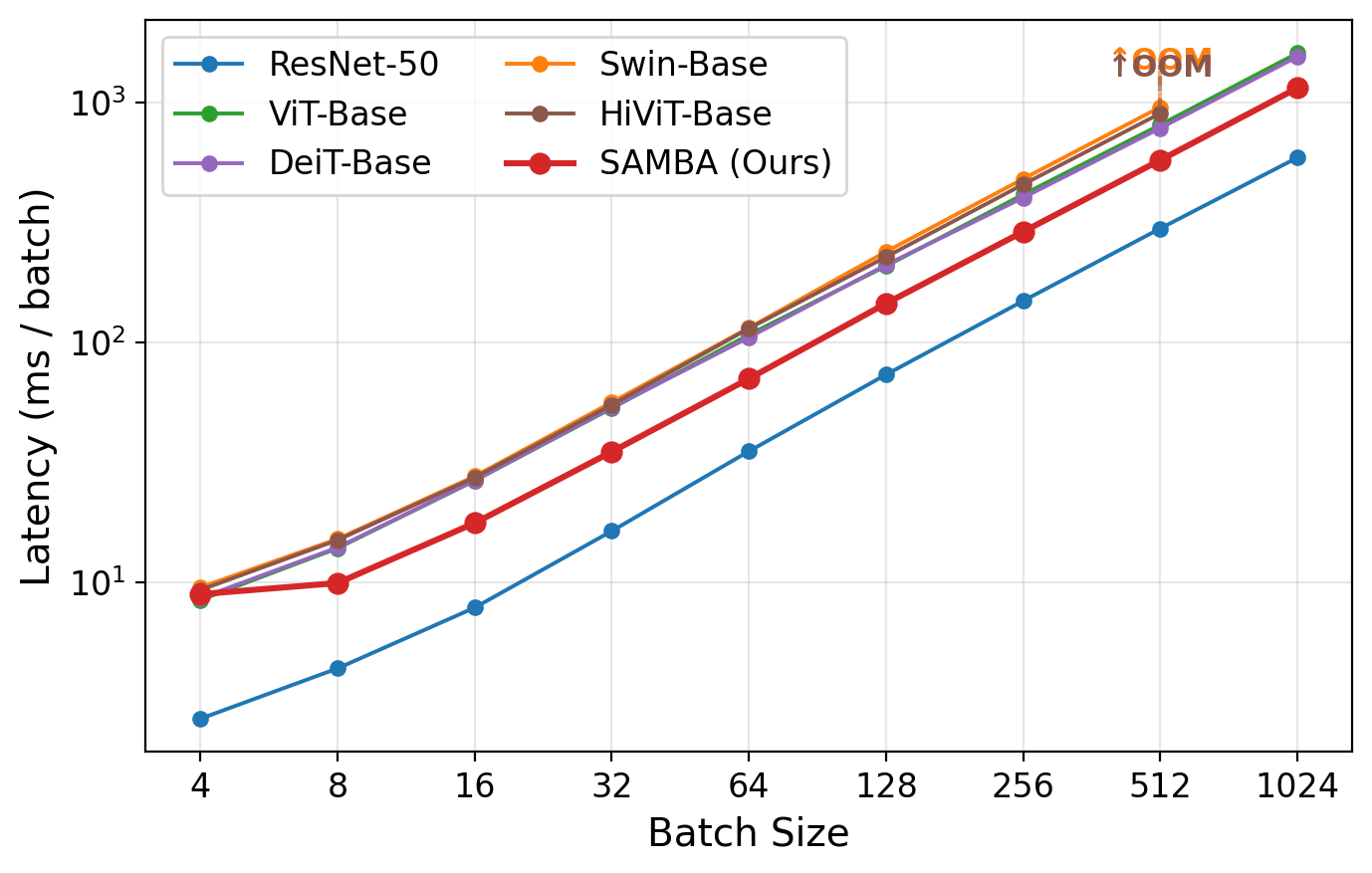}\label{subfig:b}}
	\hfill
	\subfloat[]{\includegraphics[width=0.32\textwidth]{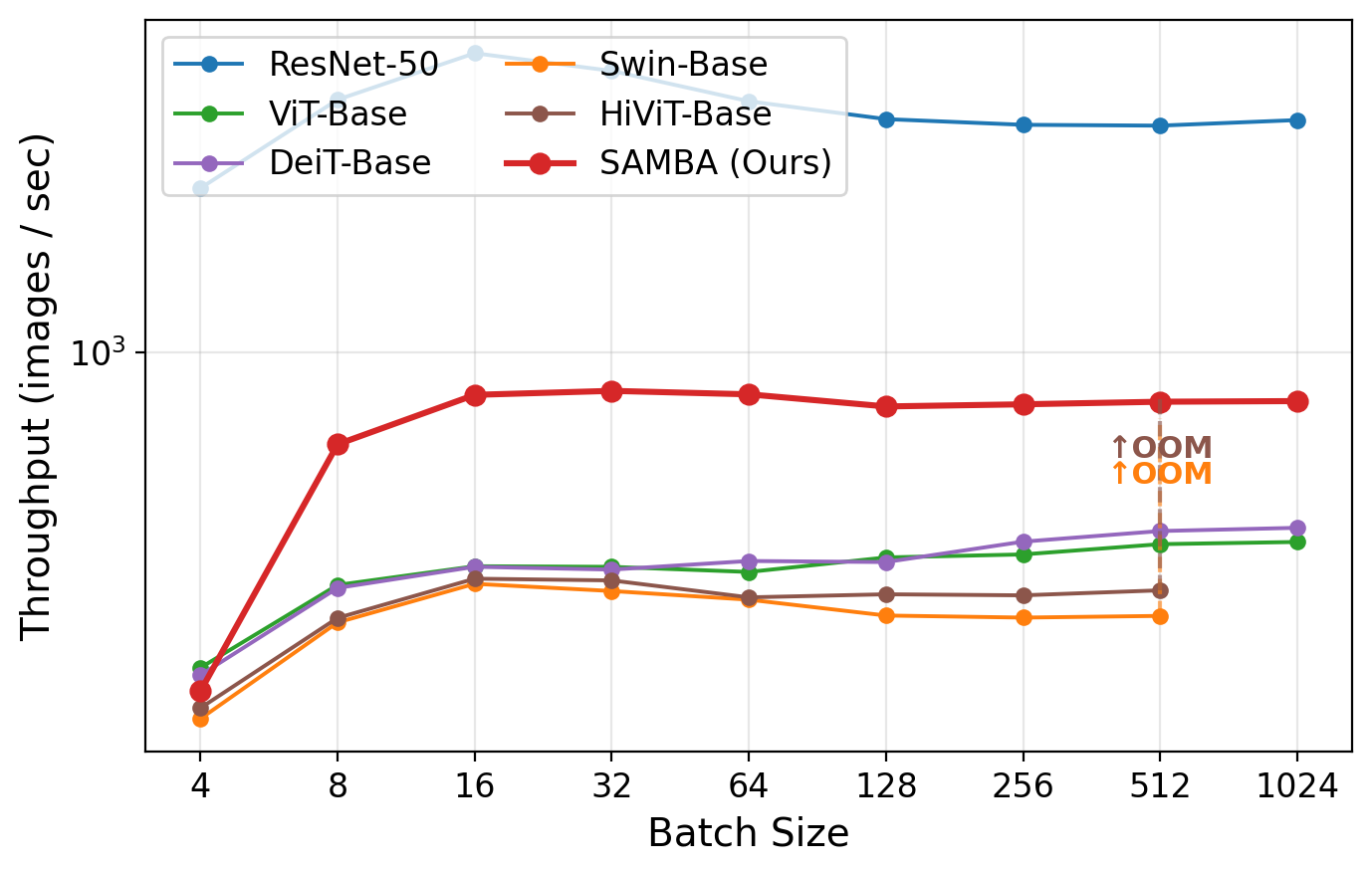}\label{subfig:c}}
	
	\vspace{1ex}
	
	\subfloat[]{\includegraphics[width=0.32\textwidth]{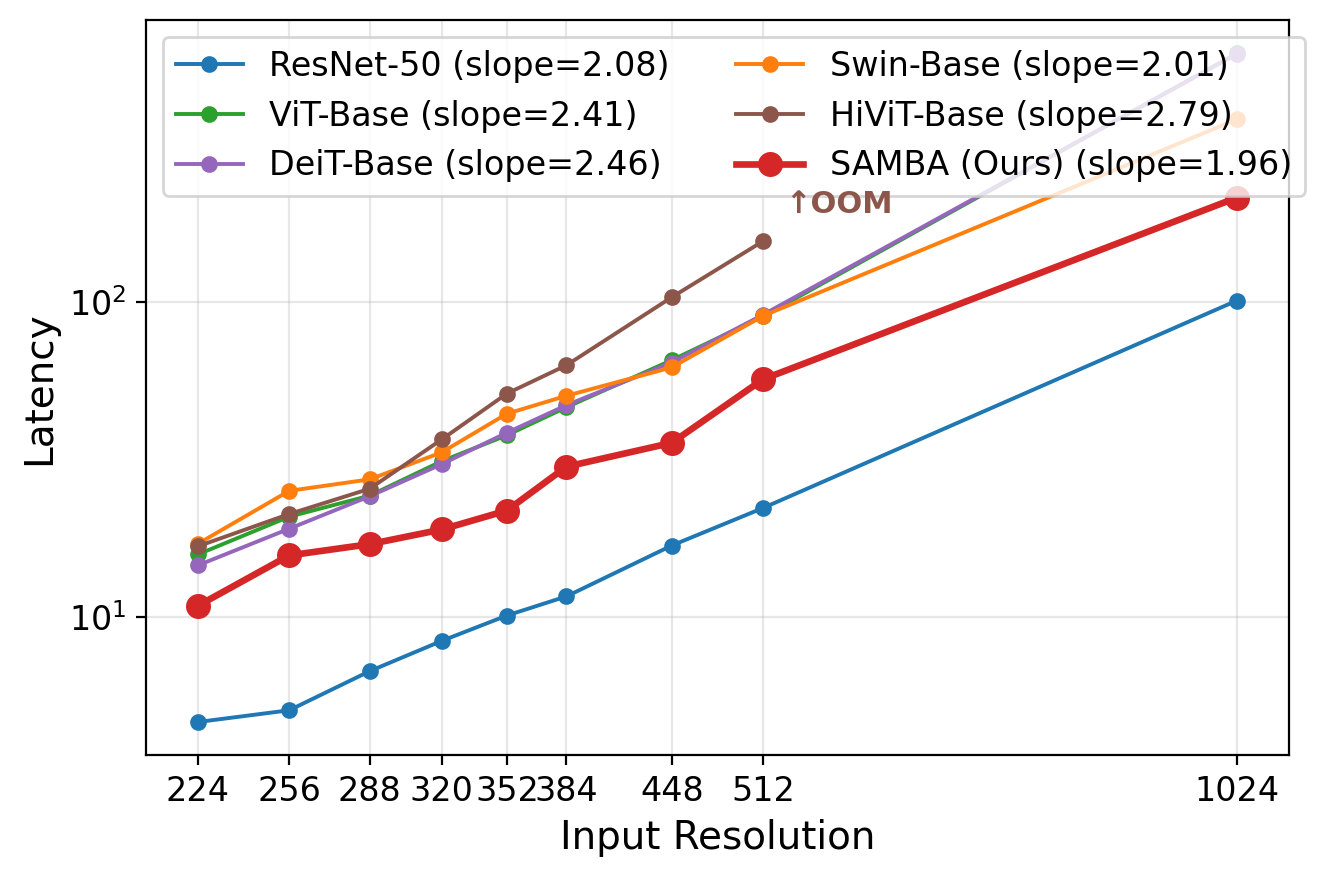}\label{subfig:d}}
	\hspace{0.04\textwidth}
	\subfloat[]{\includegraphics[width=0.32\textwidth]{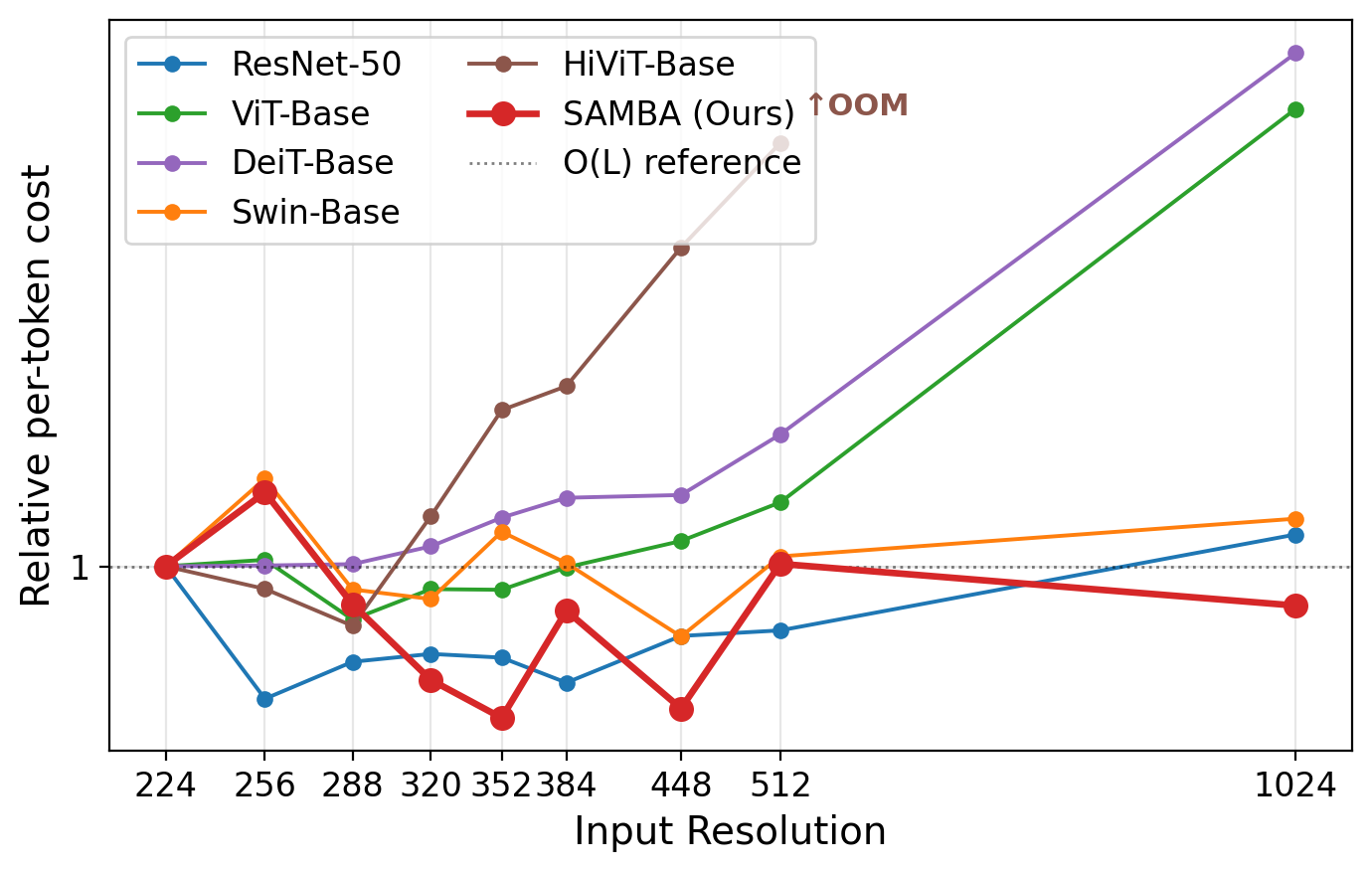}\label{subfig:e}}
	
	\caption{Comprehensive complexity and efficiency comparison between SAMBA and six representative backbones. 
		(a) Comparison of parameter count and computational complexity between the proposed model and representative architectures.
		(b) Inference latency of each model as a function of batch size, where OOM denotes out-of-memory failure of the corresponding model.
		(c) Throughput of each model as a function of batch size.
		(d) Inference latency of each model as a function of input resolution.
		(e) Computational cost growth of each model across different input resolutions.}
	\label{fig:4}
\end{figure*}

\textbf{Static Complexity Comparison.}
As depicted in \figurename~\ref{fig:4}(a), the proposed SAMBA model has approximately 27M parameters and a computational cost of around 8 GFLOPs. Compared with Transformer-based architectures, SAMBA reduces both parameter count and computational cost by over 65\%. When benchmarked against the ResNet-50 backbone, SAMBA delivers significantly superior global context modeling and feature representation capabilities with only a modest increase in model size and computational overhead.

\textbf{Impact of Batch Size on Inference Latency.}
As shown in \figurename~\ref{fig:4}(b), the per-batch inference latency of all models grows approximately linearly with batch size. For any given batch size, SAMBA consistently achieves lower latency than all Transformer baselines, with only a modest increase relative to ResNet-50. Notably, Swin-Base and HiViT-Base suffer OOM failures at large batch sizes, whereas SAMBA maintains stable operation under identical hardware constraints.

As for throughput in \figurename~\ref{fig:4}(c), SAMBA rapidly reaches saturated throughput once the batch size exceeds 16. Its peak throughput is more than twice that of ViT-Base, DeiT-Base and Swin-Base, translating to superior operational efficiency for training and inference on large-scale SAR datasets.

\textbf{Impact of Input Resolution on Computational Cost.}
As shown in \figurename~\ref{fig:4}(d) and \ref{fig:4}(e), vanilla Transformer backbones equipped with global self-attention, typified by ViT-Base and DeiT-Base, exhibit an approximately quadratic scaling of computational complexity as input resolution increases. In contrast, SAMBA consistently maintains its computational overhead close to the linear reference baseline across all tested resolutions, and even marginally outperforms ResNet-50, demonstrating a nearly linear complexity scaling behavior.

\begin{figure*}[!h]
	\centering
	\newlength{\viswidth}
	\setlength{\viswidth}{0.158\textwidth}

	\subfloat{%
		\centering
		\begin{minipage}{\viswidth}
			\centering
			\scriptsize Input\\[2pt]
			\includegraphics[width=\linewidth]{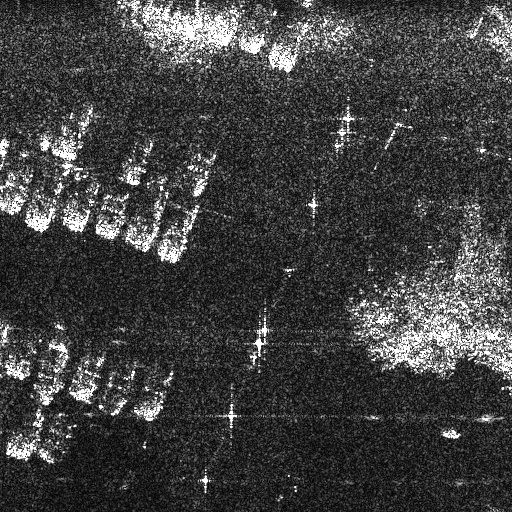}
		\end{minipage}%
	}
	\hfill
	\subfloat{%
		\centering
		\begin{minipage}{\viswidth}
			\centering
			\scriptsize Truth\\[2pt]
			\includegraphics[width=\linewidth]{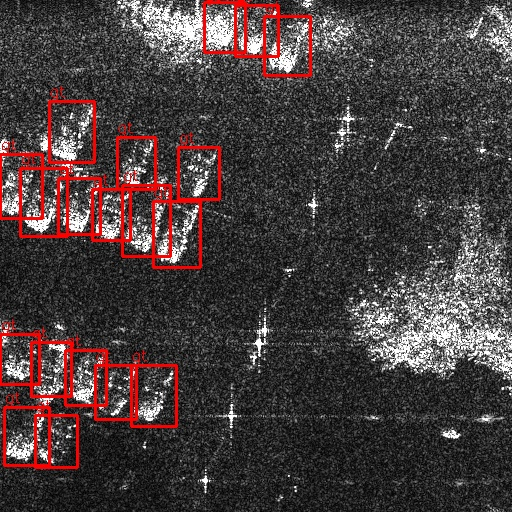}
		\end{minipage}%
	}
	\hfill
	\subfloat{%
		\centering
		\begin{minipage}{\viswidth}
			\centering
			\scriptsize Swin\\[2pt]
			\includegraphics[width=\linewidth]{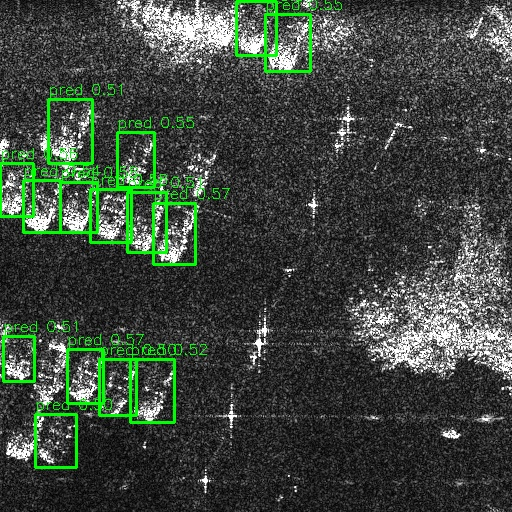}
		\end{minipage}%
	}
	\hfill
	\subfloat{%
		\centering
		\begin{minipage}{\viswidth}
			\centering
			\scriptsize ConvNeXt\\[2pt]
			\includegraphics[width=\linewidth]{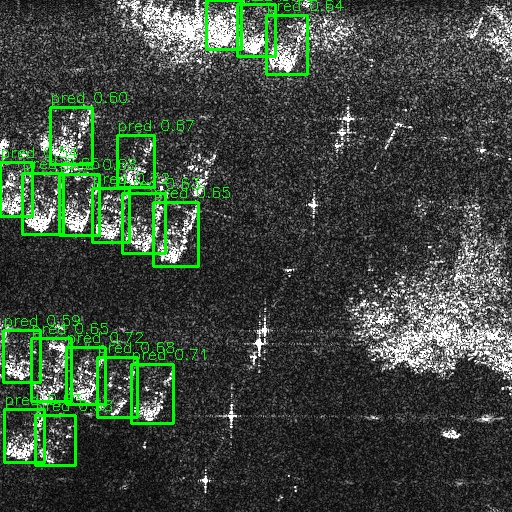}
		\end{minipage}%
	}
	\hfill
	\subfloat{%
		\centering
		\begin{minipage}{\viswidth}
			\centering
			\scriptsize ResNet-50\\[2pt]
			\includegraphics[width=\linewidth]{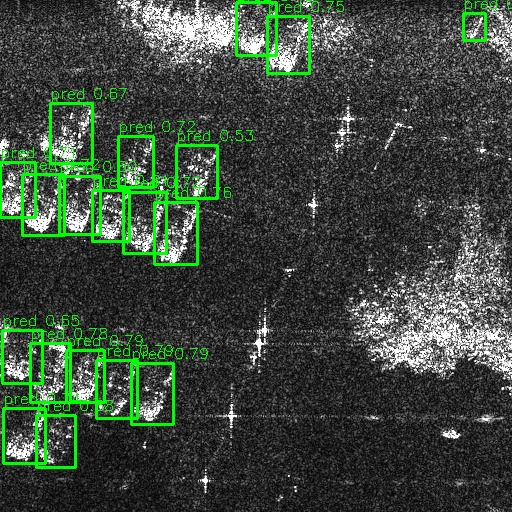}
		\end{minipage}%
	}
	\hfill
	\subfloat{%
		\centering
		\begin{minipage}{\viswidth}
			\centering
			\scriptsize SAMBA\\[2pt]
			\includegraphics[width=\linewidth]{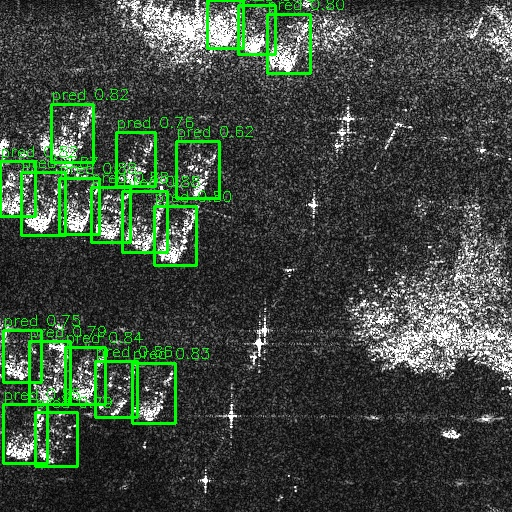}
		\end{minipage}%
	}

	\vspace{3pt}
	\subfloat{\includegraphics[width=\viswidth]{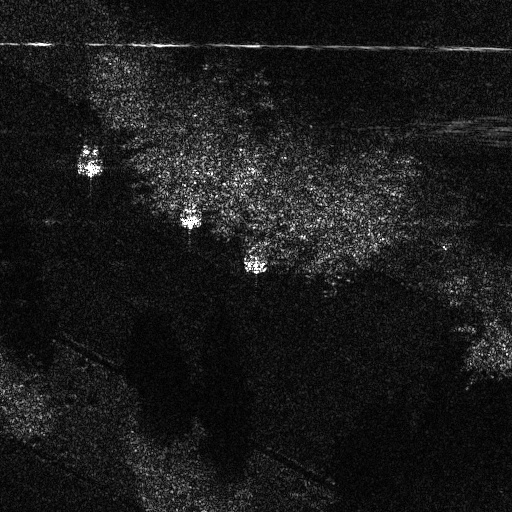}}
	\hfill
	\subfloat{\includegraphics[width=\viswidth]{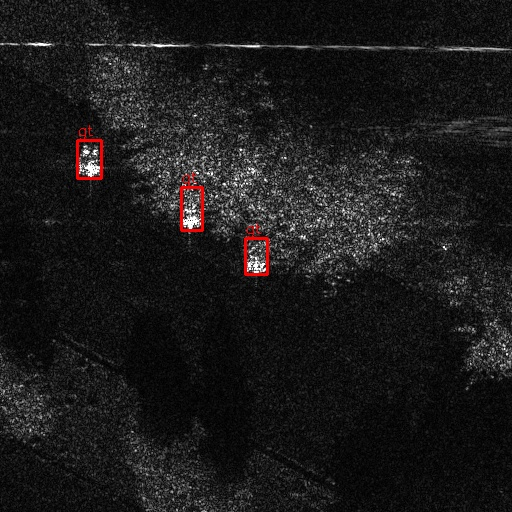}}
	\hfill
	\subfloat{\includegraphics[width=\viswidth]{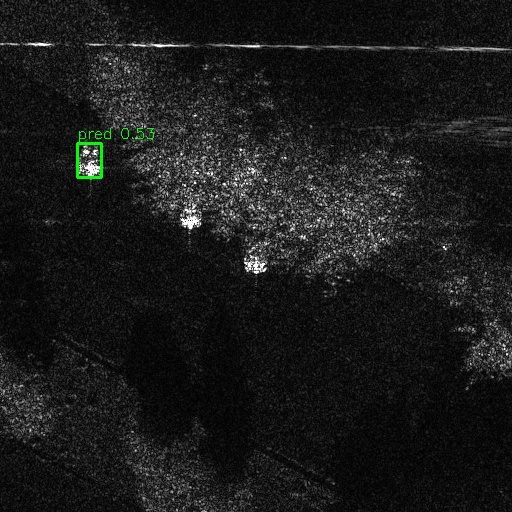}}
	\hfill
	\subfloat{\includegraphics[width=\viswidth]{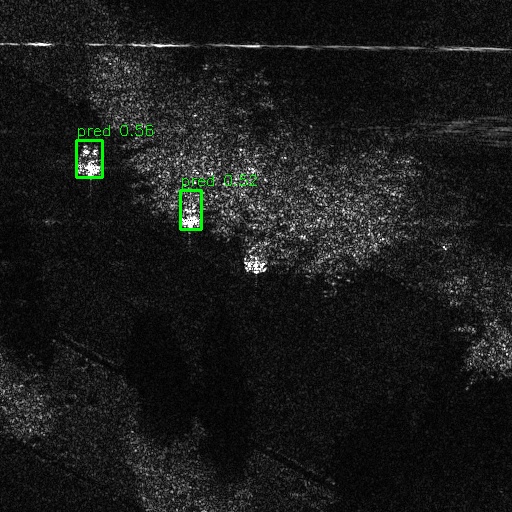}}
	\hfill
	\subfloat{\includegraphics[width=\viswidth]{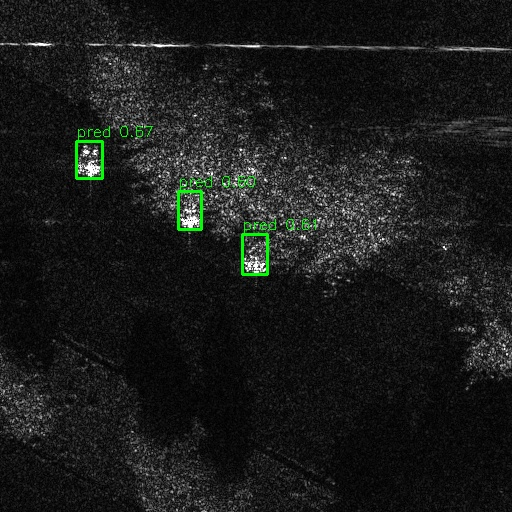}}
	\hfill
	\subfloat{\includegraphics[width=\viswidth]{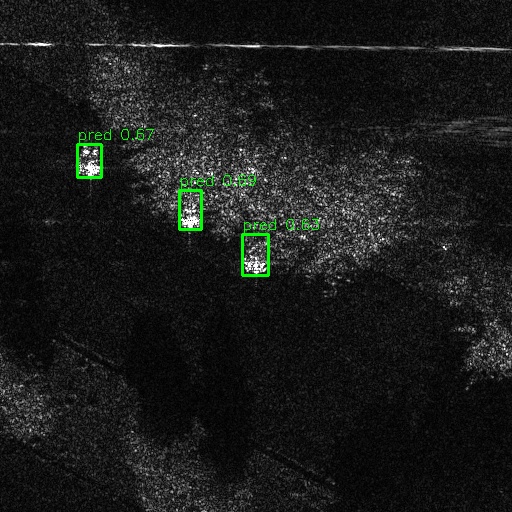}}
	
	\vspace{3pt}
	\subfloat{\includegraphics[width=\viswidth]{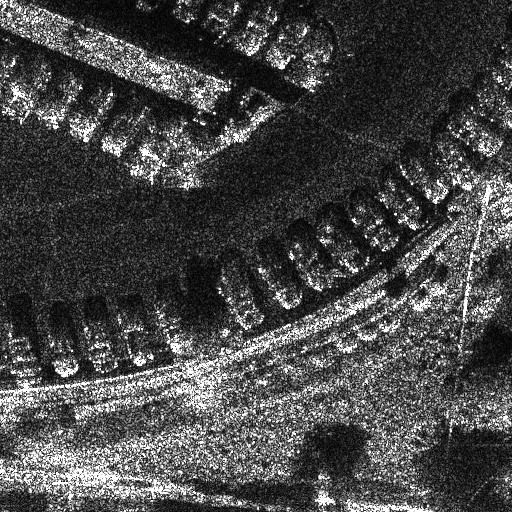}}
	\hfill
	\subfloat{\includegraphics[width=\viswidth]{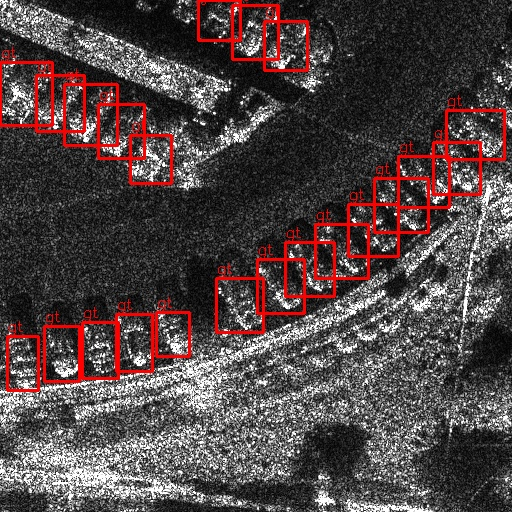}}
	\hfill
	\subfloat{\includegraphics[width=\viswidth]{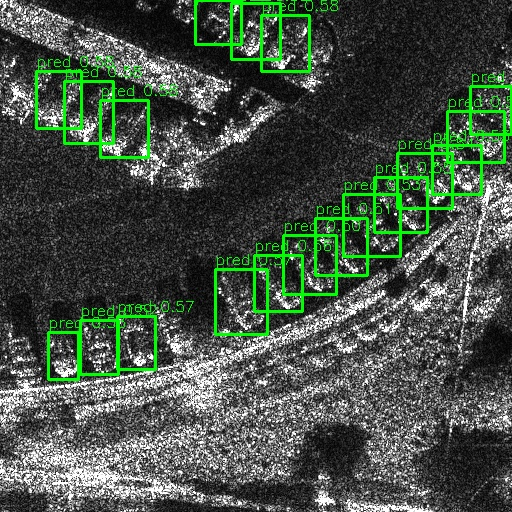}}
	\hfill
	\subfloat{\includegraphics[width=\viswidth]{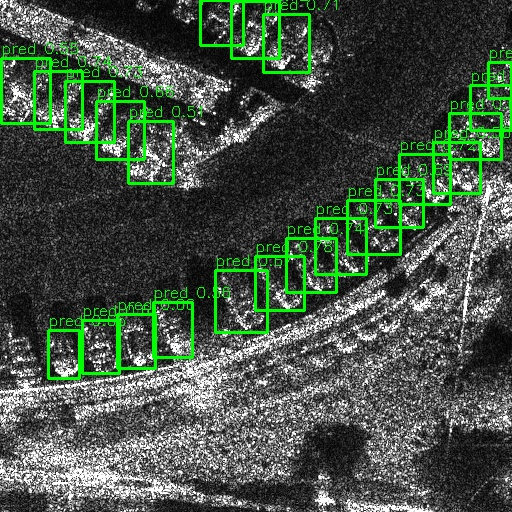}}
	\hfill
	\subfloat{\includegraphics[width=\viswidth]{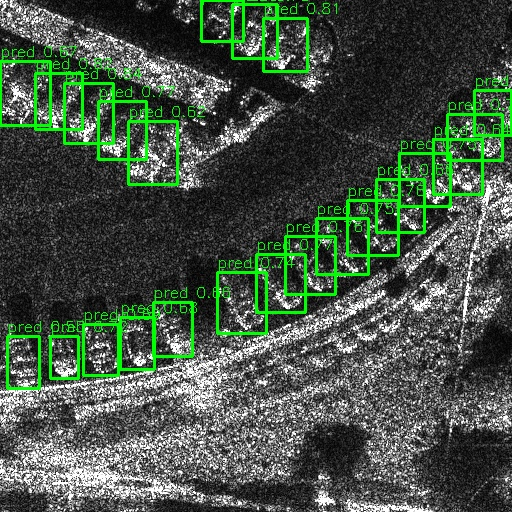}}
	\hfill
	\subfloat{\includegraphics[width=\viswidth]{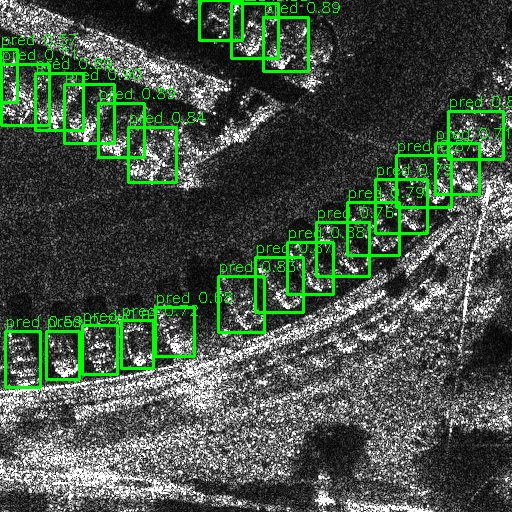}}
	
	\vspace{3pt}
	\subfloat{\includegraphics[width=\viswidth]{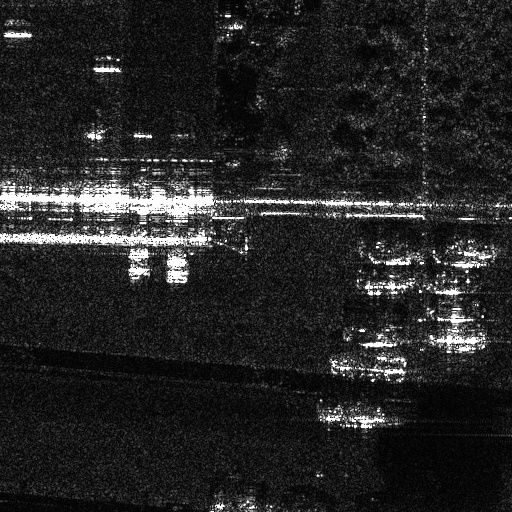}}
	\hfill
	\subfloat{\includegraphics[width=\viswidth]{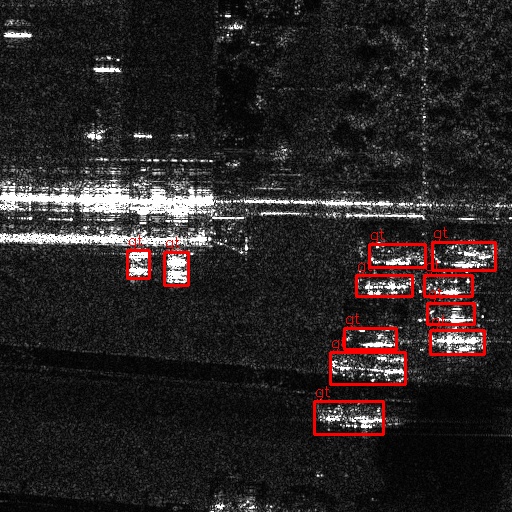}}
	\hfill
	\subfloat{\includegraphics[width=\viswidth]{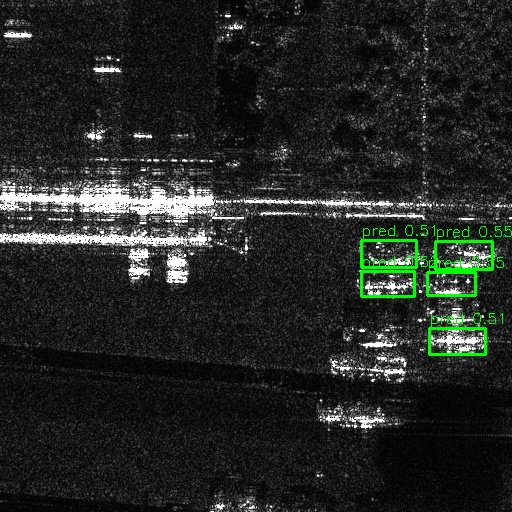}}
	\hfill
	\subfloat{\includegraphics[width=\viswidth]{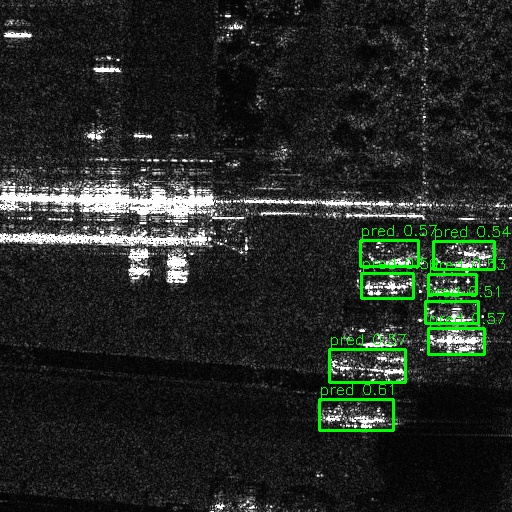}}
	\hfill
	\subfloat{\includegraphics[width=\viswidth]{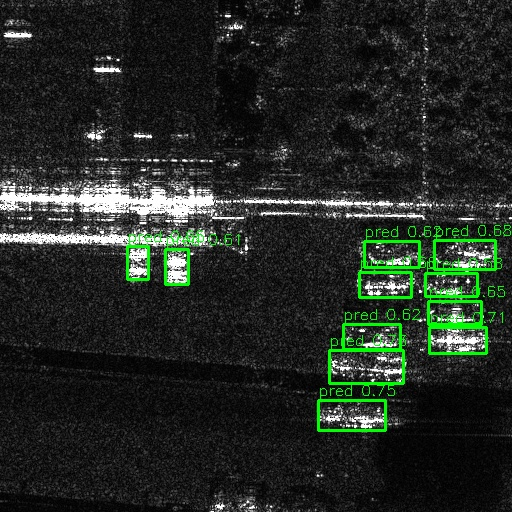}}
	\hfill
	\subfloat{\includegraphics[width=\viswidth]{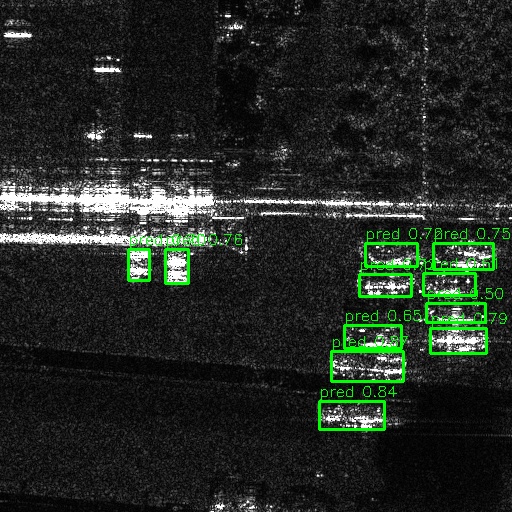}}
	
	\vspace{3pt}
	\subfloat{\includegraphics[width=\viswidth]{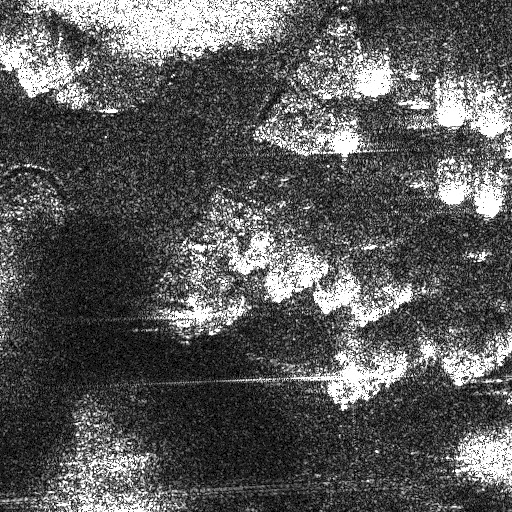}}
	\hfill
	\subfloat{\includegraphics[width=\viswidth]{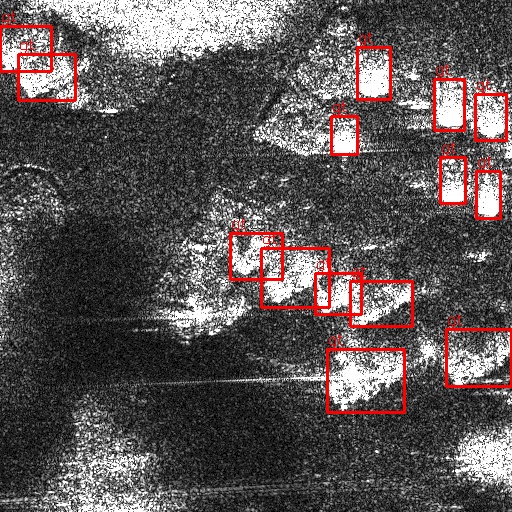}}
	\hfill
	\subfloat{\includegraphics[width=\viswidth]{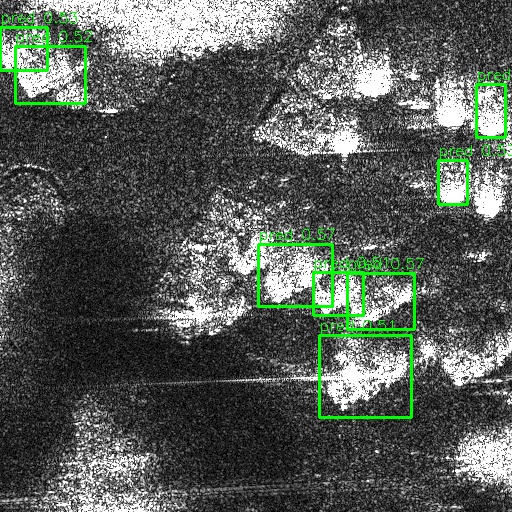}}
	\hfill
	\subfloat{\includegraphics[width=\viswidth]{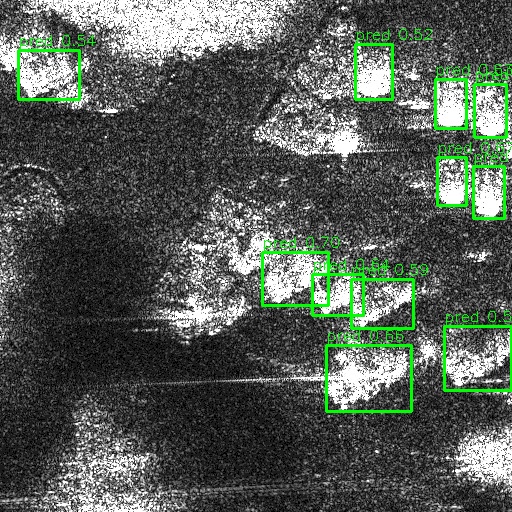}}
	\hfill
	\subfloat{\includegraphics[width=\viswidth]{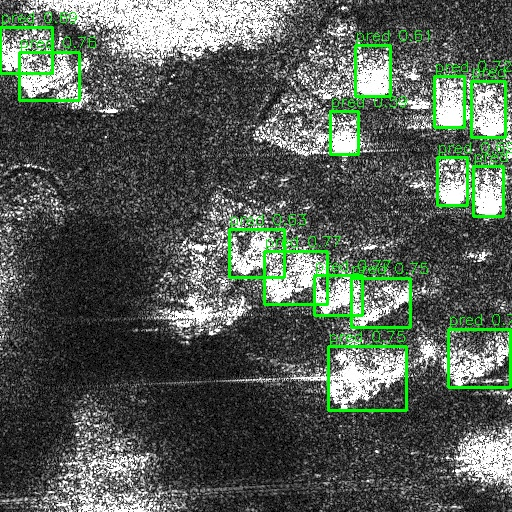}}
	\hfill
	\subfloat{\includegraphics[width=\viswidth]{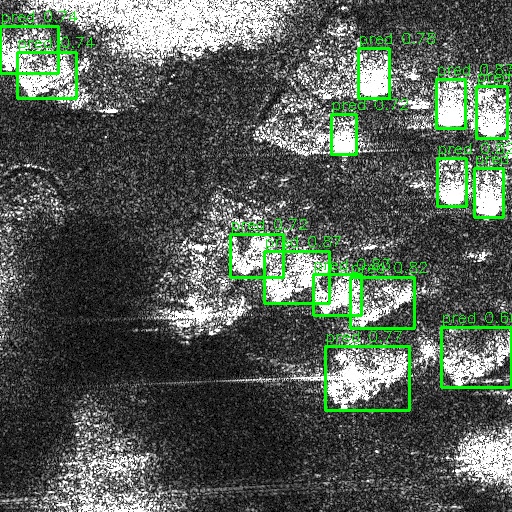}}
	
	\vspace{3pt}
	\subfloat{\includegraphics[width=\viswidth]{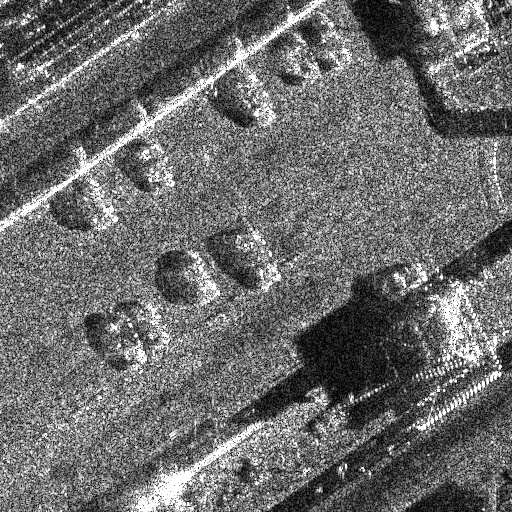}}
	\hfill
	\subfloat{\includegraphics[width=\viswidth]{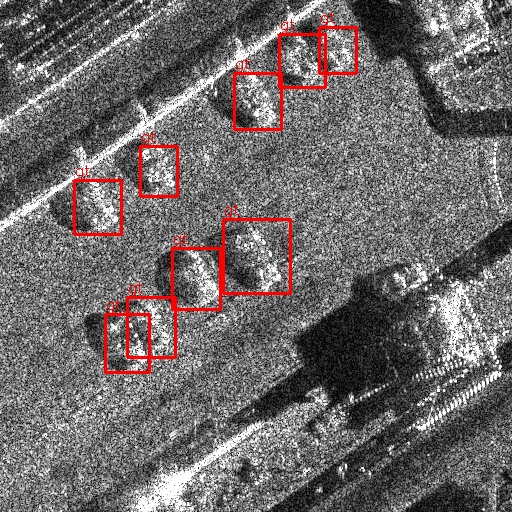}}
	\hfill
	\subfloat{\includegraphics[width=\viswidth]{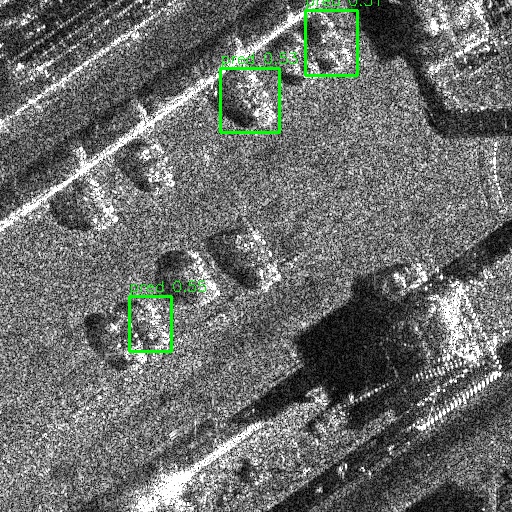}}
	\hfill
	\subfloat{\includegraphics[width=\viswidth]{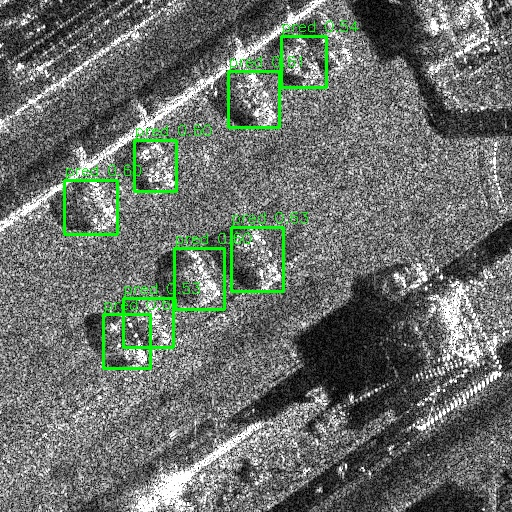}}
	\hfill
	\subfloat{\includegraphics[width=\viswidth]{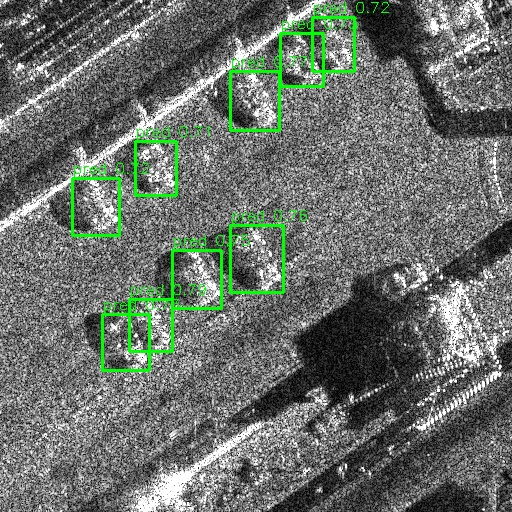}}
	\hfill
	\subfloat{\includegraphics[width=\viswidth]{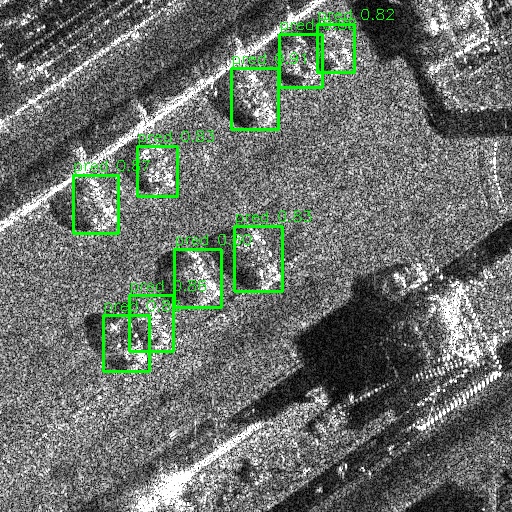}}
	
	\caption{Visualization of detection on SIVED dataset. }
	\label{fig:5}
\end{figure*}

\textbf{Target Detection Performance.}
We select representative samples covering typical target categories, multi-scale targets, densely arranged targets, and complex background clutter with inherent speckle noise. The corresponding results are shown in \figurename~\ref{fig:5}.

Compared with other baseline backbones, SAMBA achieves more accurate bounding box localization and produces fewer false positives. Benefiting from its efficient long-range dependency modeling and linear-complexity spatial modeling capability, SAMBA delivers consistently superior detection performance across diverse imaging scenarios.

\section{Conclusion}
In this work, we present SAMBA, an efficient self-supervised pre-training foundation model tailored for SAR ATR. Addressing two critical limitations of existing SAR self-supervised pre-training methods, namely the quadratic computational overhead inherent to Transformer backbones and the misalignment between generic masking methods and the unique scattering characteristics of SAR imagery, our framework embodies three core technical designs: a linear-complexity Mamba encoder with a mid-positioned class token, a three-level hierarchical SG-MAE masking strategy informed by SAR physical priors, and a lightweight SpatialMix feature interaction module.

Systematic empirical evaluations demonstrate that the lightweight Mamba backbone delivers superior performance across all pre-training configurations, while requiring substantially fewer parameters than both CNN and Transformer baselines. Our two-step cross-domain pre-training pipeline is validated as the optimal pre-training strategy, and the SG-MAE strategy further boosts few-shot transfer capability relative to the standard MAE framework. Benchmarked on seven downstream datasets spanning classification and detection tasks, the proposed method attains state-of-the-art performance on the majority of evaluation metrics, corroborating its robust generalizability across diverse SAR interpretation tasks.

For future work, we aim to extend the proposed framework to high-resolution SAR scenarios and explore its applicability to a broader range of downstream tasks, including semantic segmentation.

\bibliographystyle{IEEEtran}
\bibliography{reference}


\begin{IEEEbiography}[{\includegraphics[width=1in,height=1.25in,clip,keepaspectratio]{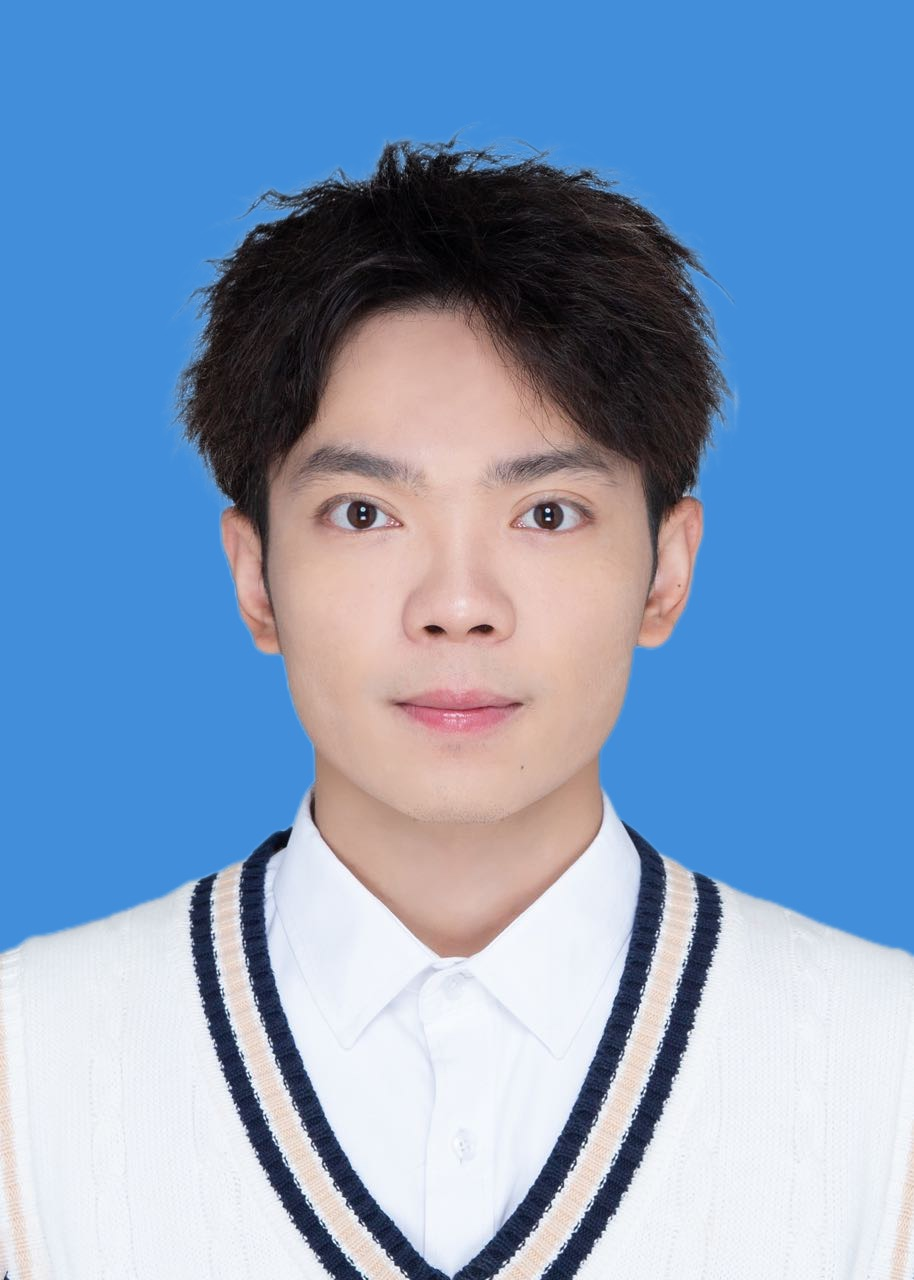}}]{Ke Wang} (Graduate Student Member, IEEE) was born in Chongqing, China, in 1999. He received the M.S. degree in Electronic Information from Air Force Engineering University, Xi'an, China, in 2025. He is currently pursuing the Ph.D. degree in Information and Communication Engineering with the College of Electronic Science and Technology, National University of Defense Technology (NUDT), Changsha, China.
	
His research interests focus on intelligent radar target recognition, intelligent radar countermeasures, and deep learning algorithms.
\end{IEEEbiography}

\begin{IEEEbiography}[{\includegraphics[width=1in,height=1.25in,clip,keepaspectratio]{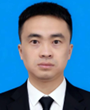}}]{Xiaoyi Pan}was born in Anhui, China, in 1986. He received the M.S. and Ph.D. degrees in information and communication engineering from the NUDT Changsha, China, in 2009 and 2014, respectively.

He is currently a Professor with NUDT. His research interests include radar countermeasure, feature extraction, and electromagnetic environment effects.
\end{IEEEbiography}

\begin{IEEEbiography}[{\includegraphics[width=1in,height=1.25in,clip,keepaspectratio]{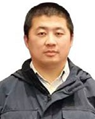}}]{Zhaoyu Gu} was born in Jiangsu, China, in 1984. He received the M.S. and Ph.D. degrees in information and communication engineering from the NUDT, Changsha, China, in 2010 and 2022, respectively.
He is currently an Associate Professor with NUDT. His fields of interest include inverse synthetic aperture radar imaging and electro-magnetic environment effects.
\end{IEEEbiography}

\begin{IEEEbiography}[{\includegraphics[width=1in,height=1.25in,clip,keepaspectratio]{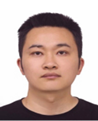}}]{Xiaofeng Ai} received the B.S. and Ph.D. degrees in information and communication engineering from the NUDT, Changsha, China, in 2005 and 2013, respectively.
	
He is currently a Researcher with NUDT. His research interests include radar target recognition, radar imaging and feature extraction.
\end{IEEEbiography}

\begin{IEEEbiography}[{\includegraphics[width=1in,height=1.25in,clip,keepaspectratio]{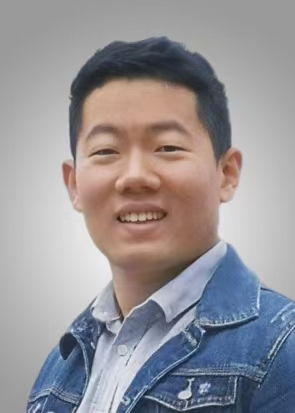}}]{Zhiming Xu} received his B.S. degree (Outstanding Graduate) from Wuhan University in 2017 and his Ph.D. degree from the NUDT in 2022. He is currently a Lecturer at NUDT. 
	
His research mainly focuses on the characteristics and recognition of polarimetric radar targets.Dr. Xu was awarded the titles of Hunan Provincial Science and Technology Innovation Young Talent and NUDT Young Elite Talent. He also serves as a Youth Editorial Board Member for multiple academic journals, a Session Chair of the ACES International Conference, and a reviewer for numerous authoritative IEEE and IET journals.In 2024, he received research grants from the Youth Fund of the National Natural Science Foundation of China and the NUDT Youth Innovation Fund, supporting his continuous innovative research in the related field.
\end{IEEEbiography}

\begin{IEEEbiography}[{\includegraphics[width=1in,height=1.25in,clip,keepaspectratio]{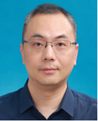}}]{Feng Zhao} was born in Jiangsu, China, in 1978. He received the B.S. degree in electronic engineering and the Ph.D. degree in information and communication engineering from the NUDT, Changsha, China, in 2001 and 2007, respectively.
	
He is currently a Professor with NUDT. His research interests include radar target recognition, radar imaging and feature extraction.
\end{IEEEbiography}

\begin{IEEEbiography}[\includegraphics{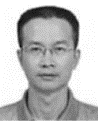}]{Shunping Xiao} received the B.E. and Ph.D. degrees in electronic engineering from the NUDT, Changsha, Hunan, China, in 1986 and 1995, respectively.
	
	He is currently a Professor with the NUDT. His research interests include radar polarimetry, SAR, signal processing, and target recognition.	
\end{IEEEbiography}

\end{document}